\documentclass{article}

\usepackage{microtype}
\usepackage{graphicx}
\usepackage{subfigure}
\usepackage{booktabs} 
\usepackage{multirow}
\usepackage{floatrow}
\usepackage{xcolor}         

\newcommand*{\rom}[1]{\textup{\uppercase\expandafter{\romannumeral#1}}}

\usepackage[pdflang={en-US}]{hyperref}

\usepackage{pdfcomment}

\usepackage[accepted]{icml2023}

\usepackage{amsmath}
\usepackage{amssymb}
\usepackage{mathtools}
\usepackage{amsthm}
\usepackage{siunitx}

\usepackage[capitalize,noabbrev]{cleveref}

\theoremstyle{plain}

\theoremstyle{definition}

\theoremstyle{remark}

\usepackage[textsize=tiny]{todonotes}

\icmltitlerunning{Neural Markov Jump Processes}

\def\*#1{\mathbf{#1}}
\def\+#1{\boldsymbol{#1}}

\begin{document}

\twocolumn[
\icmltitle{Neural Markov Jump Processes}




\begin{icmlauthorlist}
\icmlauthor{Patrick Seifner}{yyy,comp}
\icmlauthor{Rams\'es J. S\'anchez}{yyy,comp}
\end{icmlauthorlist}

\icmlaffiliation{yyy}{Lamarr Institute, Bonn, Germany.}
\icmlaffiliation{comp}{University of Bonn,  Bonn, Germany}

\icmlcorrespondingauthor{Patrick  Seifner}{seifner@cs.uni-bonn.de}
\icmlcorrespondingauthor{Rams\'es J. S\'anchez}{sanchez@bit.uni-bonn.de}

\icmlkeywords{Machine Learning, ICML}

\vskip 0.3in
]



\printAffiliationsAndNotice{}  

\begin{abstract}
Markov jump processes are continuous-time stochastic processes
with a wide range of applications in both natural and social sciences.
Despite their widespread use, inference in these models is highly non-trivial and 
typically proceeds via either Monte Carlo or expectation-maximization methods.
In this work we introduce an alternative, variational inference algorithm for Markov jump processes which relies on neural ordinary differential equations, and is trainable via back-propagation.
Our methodology learns neural, continuous-time representations of the observed data, that are used to approximate the initial distribution and \textit{time-dependent} transition probability rates of the posterior Markov jump process. 
The \textit{time-independent} rates of the prior process are in contrast trained akin to generative adversarial networks. 
We test our approach on synthetic data sampled from ground-truth Markov jump processes, experimental switching ion channel data and molecular dynamics simulations.
Source code to reproduce our experiments is available online.\footnote{\url{https://github.com/pseifner/NeuralMJP}}

\end{abstract}

\section{Introduction}

Markov Jump Processes (MJP) model time-dependent, discrete probability distributions 
%
whose future is entirely determined by knowledge of their present.
%
Given an initial distribution, the time evolution in these processes is completely characterized by the instantaneous transition probability rates between their states, in terms of which one can express quantities of interest that can be probed experimentally. 
Mean first-passage times, which provide information regarding the short-term dynamics of the process, relaxation times to stationarity and the stationary distribution itself, which concern the long-time asymptotics, or thermodynamic quantities like (stochastic) entropy production, can all be computed via the MJP transition rates.

Such a family of mathematical models is simple enough to have a wide range of approximate validity, 
%
%
which has led to its application in many fields of science.
Examples abound, and range from physics \cite{PhysRev.149.301}, chemistry \cite{gillespie77} and biology \cite{kimura1980simple} all the way up to meteorology \cite{meteoMarkovModel}, finance \cite{TURNER19893} and sociology \cite{singer1976representation}.
Indeed, there is a large class of systems whose evolution can be phenomenologically understood as the result of individual encounters between members of some population -- think of reacting chemical species or population systems, which die, mate and give birth.
Researchers can model such systems by assigning probability rates to the transitions caused by those encounters.
Likewise many large, complex systems often give rise to long-lived collective modes, 
trapped in a set of metastable states 
at some macroscopic scale of observation,
whose dynamics can also naturally be modelled in terms of transition rates between their metastable states. 
In practice, however, a judicious choice of the transition probability rates from first principles, in the large majority of these systems,
is seldom analytically tractable,
which leads researchers to resort to statistical methods in order to infer them directly from raw data.


When a continuous record of the empirical process under investigation exists, maximum likelihood estimation of the transition rates is simple and well understood \cite{Billingsley1961}. When, on the contrary, the process is observed at discrete time points only, as it is often the case, inference becomes highly non-trivial and typically proceeds via either Markov Chain Monte Carlo (MCMC) or Expectation Maximization (EM) methods.
 Yet MCMC, while accurate, does not normally scale well with data \cite{zhang2017collapsed}, whereas EM is frequently formulated to yield only point estimates of its unknown parameters \cite{opper07, koehs21} and can also struggle when faced with large datasets.

In this work we introduce an alternative, variational inference algorithm for MJP -- the Neural Markov Jump Process model (NeuralMJP) -- which relies on neural variational inference (NVI) \cite{kingma2013auto, rezende2014stochastic} and Neural Ordinary Differential Equations (NeuralODE) \cite{chen18}.
In short, our model encodes neural representations of  (noisy) data into the parameters of a master equation describing the time-evolution of the posterior process. 
Numerically solving this equation yields the instantaneous posterior distribution encoding the data.
At the same time, the model learns a global, implicit distributions over the prior MJP, through the minimization of its Kullback-Leibler divergence wrt. the posterior process.
In what follows we first review previous work on the inference of MJP (Section~\ref{sec:related_work}), and revisit some of the basics of the theory of MJP (Section~\ref{sec:background}).
Section \ref{sec:neuralMJP} introduces our model, the inference algorithm and some general details about training.
We test our methodology on four different datasets in Section \ref{sec:experiments}.
Finally, Section \ref{sec:conclusions} closes the paper with some concluding remarks about future work, while Section \ref{sec:limitations} comments on the main limitations of our approach.

\section{Related Work}
\label{sec:related_work}

The question of whether observations on a given empirical process are consistent with an underlying MJP is an old one \cite{singer1976representation}, and goes all the way back to \citet{elfving1937theorie} who formulated it as an embedding problem. 
Since then, the problem of parameter estimation for MJP has been approached from many different angles.
For example, maximum likelihood estimators for MJP transition rates were first obtained via EM by \citet{asmussen1996fitting}. Their approach was later rediscovered by \citet{bladt2005statistical} and optimized further by \citet{PhysRevE.76.066702}.
Other (maximum likelihood) formulations estimate the transition rates indirectly, by first fitting discrete-time Markov Chains to the data \cite{crommelin2006fitting, mcgibbon2015efficient}.
Likewise, Bayesian solutions to the problem of inferring MJP from noisy observations exist in many flavours. 
\citet{boys2008bayesian}, for example, proposed a MCMC algorithm for parameter inference of certain MJP classes.
\citet{fearnhead2006exact} introduced an exact sampler for Markov-modulated Poisson processes,
which was later extended and generalized, using an auxiliary-variable trick, by \citet{rao2013fast}.
The latter can however experience slow-mixing problems, and its convergence is often difficult to quantify. 
Small-variance asymptotic approaches \cite{huggins2015jump} and collapsed variational MCMC algorithms \cite{zhang2017collapsed} were recently introduced to tackle (some of) these issues.
%
%

Parallel to these efforts is a line of deterministic, variational models, the first of which was introduced by \citet{opper07}. 
Their model uses a mean-field approximation to estimate the posterior distribution of coupled MJP and is optimized via EM. 
Later \citet{cohn2010mean} applied Opper's approach to continuous-time Bayesian networks,
while \citet{wildner2019moment} partitioned the set of transitions of the MJP, to express Opper's variational objective in terms of natural moment functions.
More recently, \citet{koehs21} proposed a variational EM scheme to infer switching diffusion processes modulated by (hidden) MJPs.
Nevertheless, all these models are formulated to yield only point estimates of the MJP parameters they infer.

Finally, most neural-based approaches tackle inference of MJP mainly in specific representations.
Point processes, for example, have been modelled with recurrent \cite{du2016recurrent, mei2017neural} and NeuralODE \cite{jia2019neural, chen20} networks, both trained via maximum likelihood.
\citet{ojeda2021learning}, in contrast, leveraged recurrent, generative adversarial networks to infer queuing processes.
To the best of our knowledge, our methodology is the first, neural-based variational solution to the problem of posterior inference and parameter estimation of hidden Markov jump processes.

\begin{figure*}[t!]
    \centering
    \pdftooltip{\includegraphics[width=\textwidth]{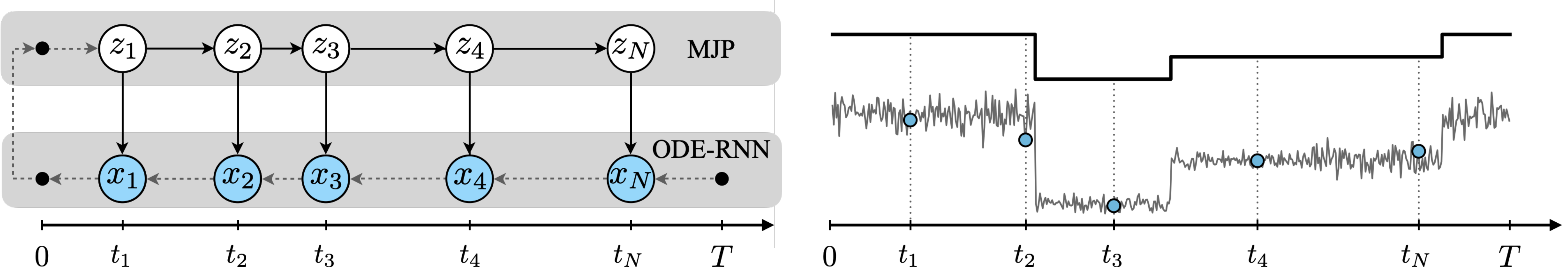}}{Left: Schema of Neural Markov Jump Process model. Two rows of circles, representing observations of a time series and their latent encodings. Arrows from latent encodings to observations indicate the generative model. Right: Schema of time series data. Sample path of a three-state, right-continuous example of a underlying switching process, deformed by gaussian noise. Small circles indicate sparse observations of noisy signal over time. 
}
    
    \caption{Neural Markov Jump Process model. Left: Graphical model and encoding-decoding procedure. The observations $\*x_1, \dots, \*x_N$ are encoded backwards in time with a ODE-RNN model (bottom block). The output ODE-RNN representations are used to condition the posterior master equation (Eq.~\ref{eq:posterior_master_eq}), which is solved forward in time (top block). The instantaneous posterior distribution is then sampled at $t_1, \dots, t_N$. The resulting states $z_1, \dots, z_N$ are decoded with the emission model. Right: Illustration of a one-dimension noisy signal (bottom) and its underlying MJP (top). The discrete observations are marked with circles at the observation times $t_1, \dots, t_N$.}
    \label{fig:model}
\end{figure*}

\section{Background}
\label{sec:background}

Let us consider a system $S$ that can be described in terms of a stochastic process $Z(t)$ taking values on some countable set ${\cal Z}$, over a finite time interval $[0, T]$.
We define the instantaneous probability rate of transiting from state $z'$ to another state $z\in {\cal Z}$ different from $z'$ as 
\begin{equation}
     f(z|z', t) = \lim_{\Delta t \rightarrow 0} \frac{1}{\Delta t} p(z, t+\Delta t| z', t),
     \label{eq:process_rate}
\end{equation}
where $p(z, t| z', t')$ is the probability to find the system in state $z$ at time $t$, given that it was in state $z'$ at time $t' < t$.
If all changes in $S$ are only due to transitions of the form of Eq.~\ref{eq:process_rate}, then $S$ is Markovian, and
one can readily show that the probability distribution describing the state of the system evolves according to the \textit{master equation}
\begin{equation}
    \dot p(z, t) = \sum_{z'\neq z} \Big( f(z|z', t)p(z', t) - f(z'|z, t)p(z, t) \Big),
    \label{eq:master_eq}
\end{equation}
where we use $\dot p(z, t)$ to denote the time derivative of $p(z, t)$. See Appendix \ref{sec:master_eq_derivation} for a brief derivation.
%
The stochastic process $Z(t)$ associated with (the solution of) this equation is made up of right-continuous, piecewise-constant paths (see Appendix~\ref{app:sampling_homog_mjp}),
which explains why one usually refers to it as a Markov jump process.

In this work we concentrate on (the inference of) \textit{homogeneous} MJP, whose transition probability rates $f$ are time-independent, 
taking values on finite state spaces of size $|{\cal Z}|=K$.
%
Note that in this case we can gather all transition rates into a $K \times K$ \textit{rate matrix} $\*F$, 
and rewrite the master equation as
%
$\dot {\*p}(t) = \*p(t) \cdot \*F$,
where $\*p(t) \in [0, 1]^{K}$ is the vector containing the time-dependent probabilities over the $K$ states in ${\cal Z}$. 
We can now formally write the solution to the master equation in terms of the matrix exponential $\*p(t) = \*p(0) \cdot \exp (\*F t)$. 

The matrix exponential representation naturally leads to the notion of \textit{relaxation times} to stationarity. Indeed, let $\lambda_1, \lambda_2, \dots, \lambda_{K}$ denote the eigenvalues of $\*F$.
If all $K$ states in $S$ are connected (i.e.~if the Markov chain associated to the process is \textit{irreducible}), the Perron-Frobenious theorem states that $0 = \lambda_1 > \text{Re}(\lambda_2) \ge \dots \ge \text{Re}(\lambda_{K})$, where $\text{Re}$ denotes real part \cite{berman1994nonnegative}.
It follows that the class probabilities $\*p(t)$ are nothing but linear combinations of the terms $1, \, \exp(-\text{Re}(\lambda_2) t), \dots, \exp(-\text{Re}(\lambda_{K}) t)$, maybe multiplied by some oscillatory functions. 
The relaxation time to stationarity is then simply given by the longest time scale in the system, namely $\text{Re}(\lambda_2)^{-1}$.
In the long-time limit, all those exponential terms vanish away and one is only left with a distribution $\*p^{*}(t)$ which, at most, oscillates among its states with time.
This limit distribution is known as the \textit{stationary distribution} of the process.
Appendix~\ref{sec:stationarity-mfpt} introduces it (properly).

The transition rates can also be used to obtain information regarding the short-TERM behaviour of the MJP. 
An important example is the \textit{mean-first passage time} $\tau_{zz'}$, which measures how long the system would take to reach some state of interest $z' \in \mathcal{Z}$, given that it started in some other state $z$, at the beginning of the process.
Appendix~\ref{sec:stationarity-mfpt} defines this time in terms of the rate matrix $\*F$.



\section{Neural Markov Jump Processes}
\label{sec:neuralMJP}

We address the general problem of inferring homogeneous MJP taking values on some countable set of size $K$, from a set of (noisy) observations on a given empirical process. 
%
%
The empirical process in question can itself lie on some countable, discrete space, like when one records the population of individuals in a given system.
It can lie on a low-dimensional, continuous space, like when one analyses one-dimensional noisy signals jumping between $K$ different mean values.
It can also lie on a high-dimensional space, like when one studies the equilibrium dynamics of the atoms in a molecule, which spend most of their time in a set of $K$ metastable, spatial configurations.
Although the posterior process associated to such noisy signals is too a MJP \cite{bishop2006pattern}, the initial distribution and rate probabilities characterizing it are complex, nonlinear functions of the entire set of observations.
This is specially true in the high-dimensional, continuous cases of conformational dynamics, typical of the molecular systems we examine below.

In what follows we use NVI methods \cite{kingma2013auto, rezende2014stochastic} to 
(i) learn neural functions encoding the discrete-time, noisy signal into the parameters of a master equation describing the time-evolution of the posterior process; 
(ii) solve, in the spirit of NeuralODE \cite{chen18}, the master equation to infer the instantaneous posterior distribution encoding the data; 
%
and (iii) learn an implicit distribution on the prior homogeneous MJP that best fits the posterior.

\subsection{Generative Model}

Let us suppose we are given a set of $N$ observations $\*x_1, \*x_2, \dots, \*x_N \in {\cal X}$ on some empirical process, recorded at (non-equidistant) times $0<t_1< t_2< \dots<t_N < T$,  
where ${\cal X}$ denotes the space of observations. 
We assume the observations are generated conditioned on
a hidden, homogeneous MJP, characterized by a master equation like Eq.~\ref{eq:master_eq}, but where the time-independent transition rates $f_{\theta}(z|z')$ are trainable functions, parametrized by some learnable parameter $\theta$.
The joint probability distribution over the prior MJP process, together with the set of $N$ observations, can then be written as
\begin{equation}
\prod_{i=1}^{N} p_{\theta}(\*x_i|z) p_{\theta}(z, t=t_i),
\label{eq:generative_model}
\end{equation}
where $p_{\theta}(z, t=t_i)$ is the (instantaneous) solution of the master equation specifying the prior process at time $t_i$, and $p_{\theta}(\*x_i|z)$ is some emission function modelling the noise in the data, with learnable parameters $\theta$. 
Let us specify each of the components in Eq.~\ref{eq:generative_model}.

\textbf{Emission Model}. The choice of the emission functions depends on the problem at hand. 
Below we shall deal with different noisy signals on $\mathbb{R}^D$, for some fixed dimension $D$. 
We therefore choose Gaussian emission models, whose mean and covariance matrices are non-linear functions of the hidden process state, modelled by neural networks.

\textbf{Prior MJP Model}. The prior MJP is characterized by its initial condition and transition probability rates. 
We are interested in learning the latter.
%
Let $\*f_{\theta} \in (\mathbb{R}^+)^{(K-1)K}$ denote the set of $(K-1)K$ independent, global parameters of the prior rate matrix.
A natural solution to the problem of learning $\*f_{\theta}$ would be to define it as a vector of randomly initialized, trainable parameters.
%
However, we have found empirically that such an approach yields fairly unstable training, which is both prone to get stuck in local minima and highly dependent on the random initialization (see Appendix~\ref{app:implicit_prior} for details).
The Bayesian alternative would be to impose and learn prior distributions on the entries of $\*f_{\theta}$ itself.
Instead we take  a more pragmatic approach. 
We define a generative neural model $\+\Phi_{\theta}: \mathbb{R}^p \rightarrow (\mathbb{R}^+)^{(K-1)K}$ that, akin to the generator in Generative Adversarial Networks \cite{NIPS2014_5ca3e9b1}, maps random $p$-dimensional vectors, sampled from a Gaussian distribution, into the transition rates $\*f_{\theta}$. That is
\begin{equation*}
    \*f_{\theta} = \+\Phi_{\theta}(\+\varepsilon), \quad \+\varepsilon \sim {\cal N}(0, \sigma),
\end{equation*}
with $\sigma$ a hyperparameter of the model.
Similar to earlier approaches \cite{mohamed2016learning, huszar2017variational, yin2018semi}, this (prior) model implicitly defines a distribution over the rate matrices of homogeneous MJP. 
Below we demonstrate that the empirical mean of this distribution is sharply peak around the parameters of ground-truth rate matrices.

\begin{algorithm}[tb]
\caption{Training NeuralMJP}
\label{alg:training}
\begin{algorithmic}
\STATE {\bfseries Requires:} Dataset $\mathcal{D}$ and time horizon $T$
\FOR{each training iteration}                       
\STATE Sample: $(t_{0:N}, \*x_{1:N}) \sim {\cal D}$ and  $\+\varepsilon \sim {\cal N}(0, 0.1)$
\vspace{0.1cm}
\STATE {\bfseries (1) Train encoder-decoder pair:}
\STATE Freeze prior parameters in $\theta$ and free the rest
\STATE Compute: $\*h_T$ = ODE-RNN($\*x_{1:N}$, $T$)
\STATE Run: \textsc{ODESolve}[$q_{\phi}(z, 0)$, $\dot q_{\phi}(z, t, \*g_{\phi})$]
\STATE Sample: $\{z_{i} \sim q_\phi(z, t_i| \*h_T) \}_{i=0}^N$ 
\STATE Backprob: $\theta, \phi \leftarrow$ Adam$(\nabla \log p_{\theta}(\*x_{1:N}|z_{0:N}))$                             
\vspace{0.1cm}
\STATE {\bfseries (2) Train prior parameters:}  
\STATE Freeze decoder parameters in $\theta$ and free the rest
\STATE Compute: $\*f_{\theta} = \+\Phi_{\theta}(\+\varepsilon)$
\STATE Backprob: $\theta \leftarrow$ Adam$(\nabla \mathcal{L}_{\text{KL}}(\*f_\theta, \*g_\phi))$
\ENDFOR
\end{algorithmic}
\end{algorithm}%

\subsection{Inference Model}

Exact inference of the generative model above is clearly intractable. 
In this section we approximate the true posterior MJP with a variational, inhomogeneous MJP solving the master equation
\begin{multline}
    \dot q_{\phi}(z, t| \*x_{1:N}) = \sum_{z'\neq z} \Big\{ g_{\phi}(z|z', t, \*x_{1:N})q_{\phi}(z', t| \*x_{1:N}) \\ - g_{\phi}(z'|z, t, \*x_{1:N})q_{\phi}(z, t|\*x_{1:N}) \Big\},
    \label{eq:posterior_master_eq}
\end{multline}
with data-dependent initial condition $q_{\phi}(z, 0|\*x_{1:N})$. 
Here $\phi$ labels the trainable parameters of the inference model, and $\*x_{1:N}$ denotes the sequence of empirical observations $\*x_{1}, \dots, \*x_{N}$.
Also note that we allow the variational approximation to capture inhomogeneous MJP, simply by making the posterior rates $g_{\phi}$ explicitly time dependent.

As can be seen already from Eq.~\ref{eq:posterior_master_eq}, our strategy consists in encoding the empirical process $ \*x_{1:N}$ into the posterior MJP indirectly, through the initial condition and transition rates defining its master equation. We define the former as
\begin{equation*}
    q_{\phi}(z, 0| \*x_{1:N}) = \+\Lambda_{\phi}(\*h_T), \quad \+\Lambda_{\phi}: \mathbb{R}^{H} \rightarrow [0, 1]^{K},
\end{equation*}
%
%
where $\*h_T \in \mathbb{R}^H$ is a neural representation encoding $\*x_{1:N}$, and $\+\Lambda_{\phi}$ is implemented with neural networks.
Accordingly, we define the $(K-1)K$ \text{time-dependent} transition rates of the posterior MJP, which we denote by $\*g_{\phi}$, as follows
\begin{equation*}
    \*g_{\phi}(t, \*x_{1:N}) = \+\Psi_{\phi}(\*h_T, t), \, \, \+\Psi_{\phi}: \mathbb{R}^H \times \mathbb{R} \rightarrow (\mathbb{R}^+)^{(K-1)K}
\end{equation*}
where $\+\Psi_{\phi}$ is implemented via neural networks too.

The only missing piece is the neural data representation $\*h_T$.
This representation can in general be computed using any sequence-processing neural network, like e.g.~a LSTM \cite{10.1162/neco.1997.9.8.1735} or Transformer \cite{vaswani2017attention} network.
We experimented somewhat with different architectures, but in the end settled on the ODE-RNN encoder of \citet{rubanova19}, which we found to be robust in all experiments without too much overhead.
For the sake of completeness, let us revisit it briefly.

\textbf{ODE-RNN Encoder}. Consider the time interval $[0, T]$ for some time horizon $T$, so that $0<t_1< \dots<t_N < T$. 
Given some initial state $\*h_0$, ODE-RNN encodes the sequence $\*x_{1:N}$ \textit{backwards in time}, starting from the time horizon $t=T$ and reaching the end of the interval at $t=0$.
ODE-RNN leverages two neural networks, namely (i) a NeuralODE to update the data representation $\*h(t)$ in between observations, and (ii) a recurrent neural network to model instantaneous updates at the observation points. 
With this procedure in mind, we define the representation $\*h_T$ encoding the complete data sequence as the output of \textit{the last} NeuralODE update, starting at $t=t_1$ and ending at $t=0$.
See \citet{rubanova19} for details.

Given $\*h_T$, we can now numerically solve the Eq.~\ref{eq:posterior_master_eq} above to obtain the instantaneous posterior distribution encoding the data at the observation times.
%
%
Thus NeuralMJP solves one (posterior) master equation for each realization of the empirical process (that is, for each time series in the dataset).
Figure \ref{fig:model} summarizes and illustrates this procedure.

\subsection{Objective Functions}
\label{sec:objective-function}

To optimize the parameter set $\{\theta, \phi \}$ of our latent variable model we maximize a variational lower bound on the logarithm of the marginal likelihood of the data. 
Following \citet{opper07}, we derive the latter using a discrete-time representation of our MJP, and taking the continuous-time limit at the end of the calculation. See Appendix~\ref{sec:elbo-derivation} for details. Our result reads

\begin{eqnarray}
\mathcal{L} & = &  \int_0^T dt \, \mathbb{E}_{q_{\phi}(z, t)} \sum_{z' \neq z} \Big\{ g_{\phi}(z'| z, t, \*x_{1:N}) -f_{\theta}(z'|z)
\nonumber \\ 
& & - g_{\phi}(z'| z, t, \*x_{1:N}) \log \frac{g_{\phi}(z'| z, t, \*x_{1:N})}{f_{\theta}(z'|z)} \Big\} \nonumber \\
& & + \sum_{i=0}^N \mathbb{E}_{q_{\phi}(z, t_i)} \left[\log p_{\theta}(\*x_i | z(t_i))\right],
\label{eq:loss}
\end{eqnarray}
where the integral is nothing but the Kullback-Leibler divergence \cite{10.2307/2236703} between prior and posterior MJP, whereas the last term is the reconstruction cost.
Let us note that the expectation values in the expression above are computed wrt. the instantaneous posterior distribution $q_{\phi}(z, t| \*x_{1:N})$. We omitted its dependence on the data in the equation above for clarity.

The training algorithm for our model is given in Algorithm~\ref{alg:training}.
Note that we train the prior model separately from the posterior (encoder) and emission (decoder) models.
See Appendix~\ref{app:2-step-prior} for our reasoning behind this. 
See also Appendix~\ref{app:index_collapse} for additional details and tricks we employ during training.
\begin{table}[t]
\caption{Inference of (ground-truth) Discrete Flashing Ratchet process. Mean and standard deviation computed with 1000 samples of the generative prior MJP model.}
\label{tab: results-dfr}
\vskip 0.15in
\begin{center}
\begin{small}
\begin{sc}
\begin{tabular}{rccc}
\toprule
             &    $V$ &     $r$ &   $b$ \\
\midrule
          Ground Truth &  $1.00  $ &   $1.00  $ &   $1.00  $ \\
\midrule
   Irregular Grid &  $1.06 \pm 0.02$ & $1.17 \pm 0.01$ & $1.14 \pm 0.02$ \\
      Shared Grid &  $0.97 \pm 0.02$ & $1.17 \pm 0.02$ & $1.17 \pm 0.01$ \\
     Regular Grid &  $0.98 \pm 0.05$ & $1.37 \pm 0.04$ & $1.39 \pm 0.04$ \\
\bottomrule
\end{tabular}
\end{sc}
\end{small}
\end{center}
\vskip -0.1in
\end{table}

\subsection{Prediction Process}
\label{sec:prediction}

NeuralMJP naturally allows for prediction of future events of the empirical process, that is, events beyond the time horizon $T$ the model was trained on.
Indeed, in order to predict the behaviour of the empirical process at some time $T+l$, for any $l>0$, one must rely on the generative process of the model, Eq.~\ref{eq:generative_model}, albeit conditioned on the past. 

Essentially, one must generate Monte Carlo samples from the posterior distribution at the end of the observation window (i.e. at $T$) and propagate the latent representations into the future, either by solving the \textit{prior master equation}, or by simulating the \textit{prior process} \cite{gillespie77} up to the  time of interest $T+l$. 
Finally, one decodes the samples from the instantaneous prior distribution back onto the space of observation via the emission model.
For completeness, we briefly revisit the classical simulation algorithm for MJP in Appendix~\ref{app:sampling_homog_mjp}.

\begin{figure}[t]
    \centering
    \pdftooltip{\includegraphics{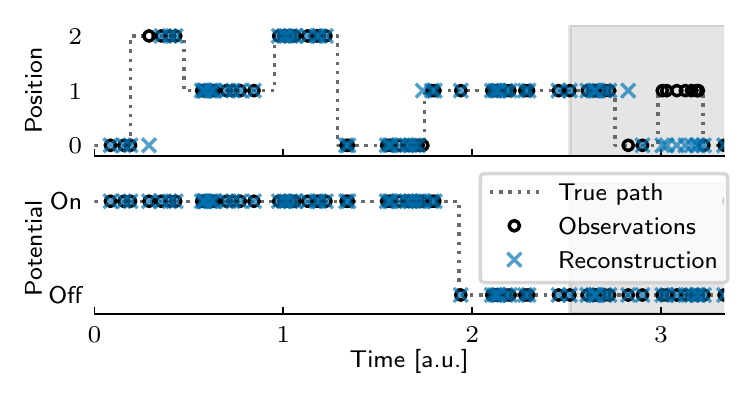}}{Discrete Flashing Ratchet sample trajectory with observations on a irregular grid and corresponding reconstructions of NeuralMJP. Position and potential of the trajectory are displayed in two separate subplots. It is split into a observation and prediction window. In the observation window, NeuralMJP reconstructs the observations at all but two observation times. In the beginning of the prediction window, the reconstruction is equally good, and only deteriorates towards the end of the prediction window. There, only the value of the potential is predicted accurately, while the predicted position $0$ deviates from the observed position $1$.}
    \caption{Sample Discrete Flashing Ratchet trajectory. The white region shows greedy samples from the posterior within the observation window. Gray region shows predictions of the model.}
    \label{fig:dfr_prediction}
\end{figure}

\section{Experiments}
\label{sec:experiments}

In this section we test our model on four different datasets of varying complexity, each characterized by different dimensionality, and corrupted by noise signals of very different nature.

We begin by studying how NeuralMJP handles irregularly sampled data, which we extract from a noiseless simulation of a simple MJP named Discrete Flashing Ratchet process. 
Next, we train the model on experimental, switching ion channel data,
that corresponds to a one-dimensional noisy signal switching among a set of mean values. 
We infer the equilibrium distribution, and relaxation and mean-first passage times of the hidden MJP, and compare against the switching diffusion model of \citet{koehs21}. 
We then use NeuralMJP to infer birth-death processes, a class of coupled (i.e.~interacting) MJPs, and compare against the classical solution of \citet{opper07}.
Finally, we infer long- and short-time properties of (one all-atom simulation of) a small molecule named alanine dipeptide, and compare against the 
%
%
neural, discrete-time variational Markov models of 
\citet{mardt17} and \citet{varolguenes19}.
Appendix~\ref{appendix:datadescription_bd} and \ref{appendix:datadescription:hybrid} contain two more experiments, in which we further compare NeuralMJP against the models of \citet{mardt17} and \citet{koehs21}, respectively.

Before diving in, let us provide some details regarding the training of NeuralMJP, which are common to all the experiments below. 

 \textbf{Experimental Details}. 
 First of all, we normalize all observation times to lie within the interval $[0,1]$ and set the time horizon $T$ to $1.1$.

To specify the posterior model, we fix the dimension $H$ of $\*h_T$ to 256 in all experiments.
Its ODE-RNN encoder uses a GRU network \cite{cho2014properties}, with a hidden dimension of 256, for the instantaneous updates, and solves a NeuralODE network, parametrized by an MLP with  $[256, 256]$ layers, backwards in time with the Runge-Kutta method, starting from $T$ with initial condition $\*h_0=0$. 
%
%
The functions $\+\Lambda_{\psi}$ and $\+\Psi_{\phi}$ are both MLPs with $[128, 128]$ and $[256,256,128]$ internal layers, respectively.
The posterior master equation (Eq.~\ref{eq:posterior_master_eq}) is solved with the adaptive step Dormand–Prince method. 
The Kullback-Leibler divergence in Eq.~\ref{eq:loss} is approximated via Gaussian quadrature with $200$ points. 
Next, and regarding the generative prior, we set the dimension $p$ of the random input vector $\+\varepsilon$ to $64$, and model $\+\Phi_{\theta}$ with a single hidden layer of 64 units.
The emission model, when used, is set to a Gaussian model, whose mean and variance are modelled with and MLP of $[128,128]$ layers.

Finally, all parameters are optimized using Adam \cite{kingma2014adam}, with learning rates varying from $\num{1e-3}$ to $\num{1e-4}$, and possible learning rate annealing. 
We provide additional details on datasets, architectures and training in Appendix \ref{app:reproducibility_and_experiments}.
We also report on training times and resource consumption of NeuralMJP in Appendix~\ref{app:training_times}.

\begin{figure}[t]
    \centering
    \pdftooltip{\includegraphics{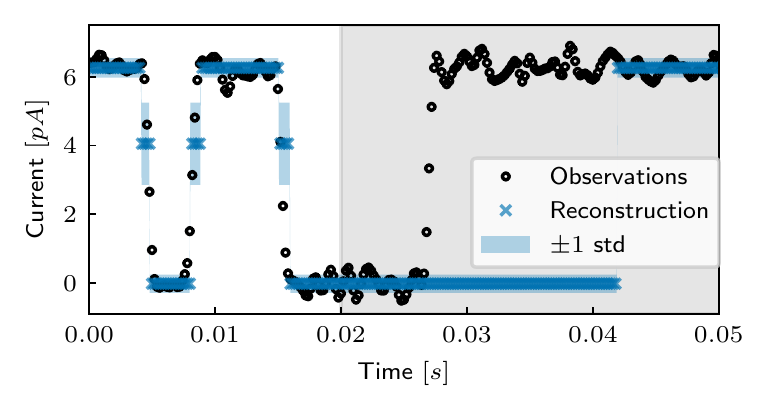}}{
    Trajectory from the Switching Ion Channel dataset with corresponding reconstructions from NeuralMJP. The measured current fluctuates continuously between $6$pA and almost $0$pA. The trajectory is split into a observation and prediction window. In the observation window, the reconstruction follows the observations closely and a intermediate state at $4$pA is resolved. In the prediction window, the trajectory transitions only once from around $0$pA to $6$pA, which the model also predicts, although later than the observed trajectory.
    }
    \caption{Sample trajectory from Switching Ion Channel dataset. The white region shows mean reconstruction values, conditioned on posterior samples within the observation window. Gray region shows predictions of the model.}
    \label{fig:ion_prediction}
\end{figure}

\subsection{Discrete Flashing Ratchet}
\label{sec:dfr}
Before testing our methodology on real-world inference problems, we evaluate the ability of the model to infer ground-truth MJPs in a controlled experiment. 
To this end, we consider the Discrete Flashing Ratchet (DFR) process, introduced by \citet{ajdari1992mouvement}.

The DFR process is a six-state MJP, which models the position of a particle that is subject to a periodic, asymmetric potential,
and is in contact with a thermal bath (here, of unit temperature).
The particle has three available states and we denote them with $i \in \{0,1,2\}$.
The potential has two states, and we denote them with ON or OFF. 
The transition rates of the DFR process are defined as follows \cite{roldan10}
\begin{eqnarray}
f(j, \text{ON} \mid i,  \text{ON}) & = & \exp\left(- \frac{V}{2}(j - i)\right), \quad \text{for} \,\, \,  i \neq j, \nonumber \\
f(j, \text{OFF} \mid i, \text{OFF}) &= &b, \quad \text{for} \,\, \, i \neq j, \nonumber \\
f(i, \text{OFF} \mid i, \text{ON}) &= &f(i, \text{ON} \mid i, \text{OFF})= r, \nonumber 
\end{eqnarray}
where $V$, $r$ and $b$ are parameters of the model. 
We set them all to $1$ in what follows and simulate a set of trajectories from the DFR process thus defined.
%

With such a simple process at hand we set out to investigate how NeuralMJP handles observations recorded at non-equidistant times.
%
%
To do this, we sub-sample three different datasets from the set of simulated DFR trajectories: \textsc{(i) Irregular Grid}, in which each realization (i.e. each trajectory) has a different, irregular grid; \textsc{(ii) Shared Grid}, in which all realizations share a fixed, irregular grid; and \textsc{(iii) Regular Grid}, in which all realizations have a fixed and regular grid. 
We do not corrupt these datasets with any noise model.
Accordingly, we trained a NeuralMJP without emission model and use a cross-entropy loss as reconstruction cost.
We also only allow transition rates in the prior MJP model which are consistent with the ground truth.
The posterior is however left unconstrained. 
Additional details about the data simulation process, data preprocessing, hyperparameters and training procedure can be found in Appendix~\ref{app:dfr}.

\begin{table}
\caption{Stationary distribution for the switching ion channel process. We report the distributions inferred by NeuralMJP when trained on both the one-second window (\textsc{1 sec}) and the complete dataset (\textsc{Full}).}
\label{tab:ion-equilibrium}
\vskip 0.15in
\begin{center}
\begin{small}
\begin{sc}
\begin{tabular}{rccc}
\toprule
                                           &             Bottom &          Middle &             Top \\
\midrule
      \citet{koehs21} &        $0.17961  $ &     $0.14987  $ &     $0.67052  $ \\
        
\midrule
                     NeuralMJP (1 sec) &    $0.17672$ & $0.09472 $ & $0.72856$ \\
                     NeuralMJP (full) &     $0.19631$ & $0.07153 $ & $0.73217$ \\
\bottomrule
\end{tabular}
\end{sc}
\end{small}
\end{center}
\vskip -0.1in
\end{table}

\begin{table*}
\caption{Inference of (ground-truth) Lotka-Volterra process against the baselines. Mean and standard deviation values are computed with 1000 samples of the generative prior MJP model. All values are of the order of $\num{e-04}$.}
\label{tab:lv-results}
\vskip 0.15in
\begin{center}
\begin{small}
\begin{sc}
\begin{tabular}{rcccc}
\toprule
                           &                $\alpha$ &                 $\beta$ &                $\delta$ &                $\gamma$ \\
\midrule
                        Ground Truth &      $5  $ &      $1 $ &      $1  $ &      $5 $ \\
 \citet{opper07} &      $ 13.5  $ &      $2.32  $ &      $1.78  $ &      $15.7  $ \\
\midrule
NeuralMJP & $4.5 \pm 0.3$ & $ 1.00 \pm 0.08$ & $ 0.76 \pm 0.08$ & $ 3.5 \pm 0.2$ \\
\bottomrule
\end{tabular}
\end{sc}
\end{small}
\end{center}
\vskip -0.1in
\end{table*}

\textbf{Results}. Table~\ref{tab: results-dfr} shows the values inferred by NeuralMJP from our three artificial datasets. 
Specifically, we report the empirical mean and standard deviation computed from 1000 samples of the generative prior MJP.
Note how all results are remarkably close to (and sharply peak around) the ground-truth values, which indicates that NeuralMJP can successfully infer MJP from sequential data. 
Interestingly, the model performs (in average) best on the \textsc{Irregular Grid}, arguably because the dataset contains more information about the underlying process.
All in all we conclude that NeuralMJP handles irregularly sampled data well.

Figure~\ref{fig:dfr_prediction} shows a sample trajectory from the DFR process, together with a NeuralMJP trajectory. 
The white region corresponds to the observation window and displays greedy samples from the posterior model. The gray region corresponds to future (unseen) data and shows predictions of the (prior) MJP model.
The posterior process clearly models the data well. The prediction process captures correctly some of the jumps too.

For completeness, we have also trained 4 additional models, with different initializations, on all three datasets.
We report the mean and std. of our results in Table~\ref{tab:app-additonal-dfr} of Appendix~\ref{app:dfr}. 
Our conclusions remain unaltered.


\subsection{Switching Ion Channel Data}
\label{sec:iondata}

In this section we study the conformational switching of the viral potassium channel $\text{Kcv}_{\tiny \text{MT325}}$, which is known to switch between three different configurations \cite{gazzarrini2006chlorella}.
Markov models are a natural choice to describe the switching processes of ion channels \cite{rauh2017identification, rauh2018site} and our methodology provides a scalable alternative to directly infer both, long- and short-term dynamic properties of the switching process, without the need to resort to frequentist approaches or introduce artificial lag time scales.

Let us assume that underlying the empirical switching process is a three-state MJP. We denote its states as \textsc{Top}, \textsc{Middle} and \textsc{Bottom}.
Each state is characterized by its ion permeability, which can be experimentally measured by applying a voltage drop across the cell membrane and recording the ion flow. 
Here we analyse a current signal which was generated by applying a voltage drop of 140 mV, and recorded at a frequency of 5 kHz over a period of about 34 seconds.
The dataset was made available to us via private communication (see the acknowledgements below).
Figure~\ref{fig:ion_prediction} illustrates a snapshot of the current signal.
One can visually identify the \textsc{Top} and \textsc{Bottom} states, with small, noisy oscillations about them.

We train NeuralMJP with a Gaussian emission model on (i) a one-second window of the data (5000 observations) and (ii) the complete dataset (without the first burn-in second, which amounts to about 167000 observations). 
See Appendix~\ref{app:ion_channel} for details.
We compare against the variational, switching diffusion model of \citet{koehs21}. 
%
%
Note that they trained their model on the one-second window of the data (i.e. the same window we consider).

\textbf{Results}. We report in Table~\ref{tab:ion-equilibrium} the stationary distribution (of the hidden process) inferred by each model. 
For NeuralMJP, this is done by sampling the prior rate matrix 1000 times, computing its corresponding stationary distribution and averaging.
The results agree well. 
Note in particular that the \textsc{Middle} state is the least likely, according to both models.
Very importantly, the distributions inferred by NeuralMJP in both datasets agree well too.

As regards the short-term dynamics, we report in Table~\ref{tab:ion_mfpt_appendix} of Appendix~\ref{app:ion_channel} the estimated mean-first passage times.
NeuralMJP infers shorter transition times to and from the \textsc{Middle} state, as compared to the baseline, which implies that NeuralMJP resolves transitions between \textsc{Top} and \textsc{Bottom} states via fast transitions to the \textsc{Middle} state.
This feature is consistent in both short and long datasets.
In contrast, the model of \citet{koehs21} resolves $\text{\textsc{Top}} \Leftrightarrow \text{\textsc{Bottom}}$ transitions that do not pass through the \textsc{Middle} state.

Figure~\ref{fig:ion_prediction} shows the NeuralMJP reconstruction of a test trajectory, both within and without the observation window.
The model clearly fits the data well.
Figure~\ref{fig:ion_channel_prediction_comparison} in the Appendix additionally compares the prediction capabilities of NeuralMJP against the model of \citet{koehs21}. The performance of both models is similar. 

\subsection{A Birth-Death Process: Lotka-Volterra}
\label{sec:LV}

In this section we leverage NeuralMJP to infer coupled MJP.
Let us consider the Lotka-Volterra process (LV), the stereotypical Birth-Death Process, as our ground-truth, coupled MJP.
The LV process takes values on $\mathbb{N}\times\mathbb{N}$ and models the population levels of prey and predator species as they interact over time.
The process is defined in terms of four positive parameters ($\alpha$, $\beta$, $\gamma$ and $\delta$), which control the strength of the species interaction, as well as the birth and death rate of prey and predator, respectively. 
In our experiments, we fix the tuple ($\alpha$, $\beta$, $\gamma$, $\delta$) to $(5, 1, 1, 5)\times 10^{-4}$ and simulate a set of trajectories from the LV process.
The task is to infer these four parameters from a dataset of noisy observations on the LV process.
See Appendix~\ref{appendix:datadescription:lv} for details.

%

\textbf{Results}. To handle the coupling in the LV process we follow \citet{opper07} and use their mean-field approximation.
Table~\ref{tab:lv-results} contains the inferred parameters of the LV process against the point estimates reported by \citet{opper07}.
Our inferred parameters are clearly closer to (and sharply peak around) the ground-truth values as compared to the baseline.
Nevertheless, let us remark that the latter was trained on a significantly smaller dataset. 
For the sake of a fairer comparison, we trained NeuralMJP on the same small dataset used by the baseline. Appendix~\ref{appendix:datadescription:lv} contains our results, which show our model outperforms the baseline in this case too. 
%
%
These findings indicate our model can successfully implement the mean-field trick and infer coupled MJP from noisy data.

Figures~\ref{fig:lv_prediction} and \ref{fig:lv_prediction-2} in the Appendix show the NeuralMJP reconstructions of LV trajectories, for models trained on both large and small datasets.

%

\subsection{Alanine Dipeptide}
\label{sec:adp}

\begin{table}[b]
\caption{Stationary distributions for six-state Markov models of ADP. VAMPnets results were extracted from \citet{mardt17}. The states are ordered such that the protein conformations associated to a given state are comparable in both models. }
\label{tab:adp_stationary}
\vskip 0.15in
\begin{center}
\begin{small}
\begin{sc}
\begin{tabular}{rcccccc}
        \toprule
        & \multicolumn{6}{c}{Probability per State} \\
        & $\rom{1}$ & $\rom{2}$ & $\rom{3}$ & $\rom{4}$ & $\rom{5}$ & $\rom{6}$ \\
\midrule
        VAMPnets  & $0.30$ & $0.24$ & $0.20$ & $0.15$ & $0.11$ & $0.01$ \\
    \midrule
        NeuralMJP    & $0.30$ & $0.31$ & $0.23$ & $0.10$ & $0.05$ & $0.01$ \\
        \bottomrule
\end{tabular}
\end{sc}
\end{small}
\end{center}
\vskip -0.1in
\end{table}

Alanine dipeptide (ADP) is a well-studied, $22$-atom molecule that serves as a benchmark system for molecular dynamics (MD) simulations and computational biology \cite{rossky1979solvation}.
When ADP is in equilibrium, its energy surface consists of a multitude of local minima. 
%
However, 
the structures of the protein associated to these minima depend mainly on the relative configuration of two central backbone atom sequences \cite{ramachandran63}. 
This relative configuration can be represented by two Ramachandran angles ($\psi$ and $\phi$), which are dihedral angles of the two atom sequences (see e.g. Fig.~\ref{fig:adp_heavy_atoms} in the Appendix). 
Projection to these angles reveal metastable conformations of the protein, each containing several local minima \cite{mironov19}.
\begin{figure}
    \centering
    \pdftooltip{\includegraphics{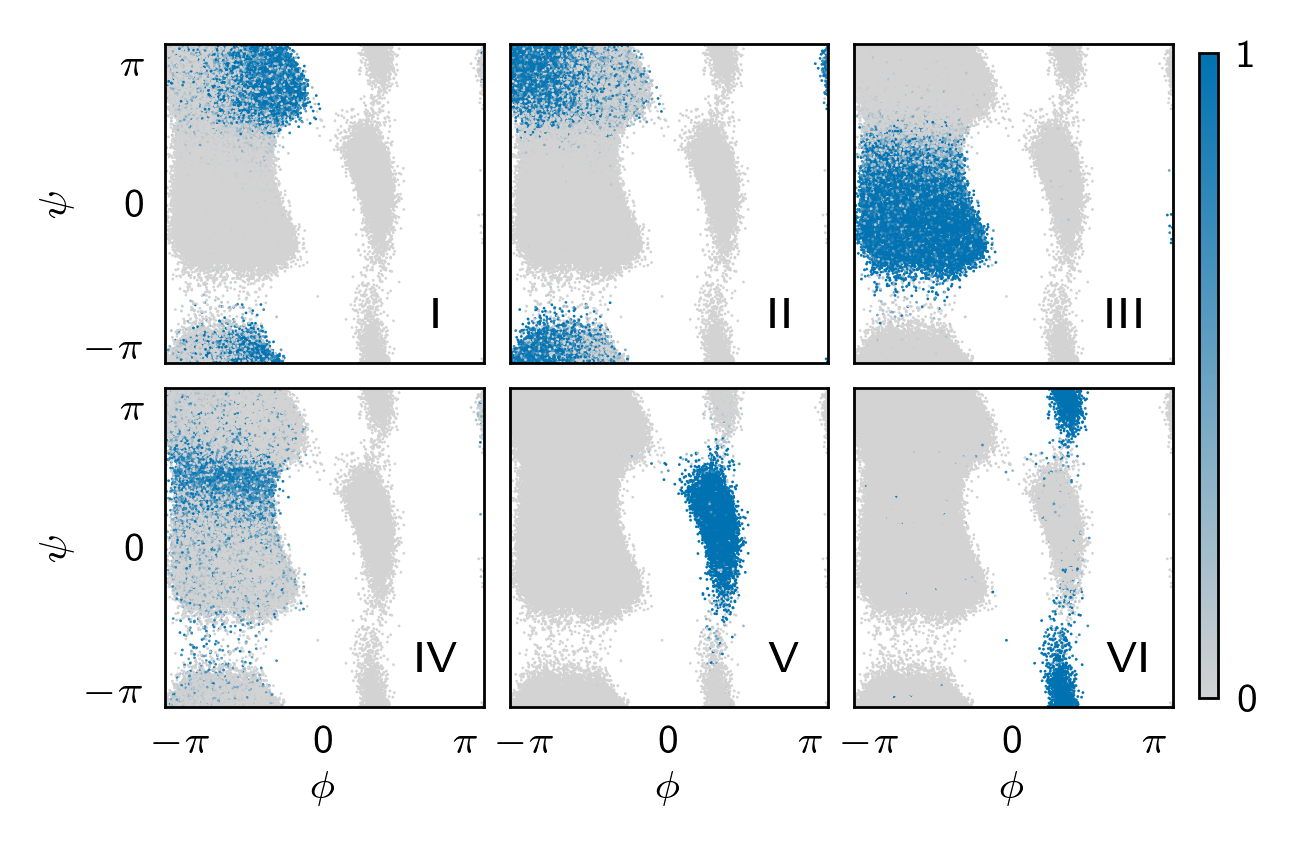}}{
    State separation of a six-state NeuralMJP model trained on Alanine Dipeptide Protein data. There are six subplots, arranged in a two-by-three grid, each containing all observations of the dataset represented in Ramachandran angles. The coloring of a datapoint in a subplot indicates the posterior probability mass that NeuralMJP puts on the datapoint for each state. There are six distinct clusters. At their edges, the cluster blend together, indicating a fuzzy separation of states learned by NeuralMJP.
    }
    \caption{Alaine Dipeptide Protein six-state separation. For each observation, the intensity of blue in subplot $i= \rom{1}, \dots, \rom{6}$ indicates the posterior probability mass for state $i$ at that observation. }
    \label{fig:adp_6_states}
\end{figure}

While the conformations themselves have been studied extensively in the past \cite{hermans11}, their dynamics are still inadequately understood.
Trajectory analysis of MD simulations is one typical approach to investigate the slow dynamics, but is generally plagued with sampling problems, specially when dealing with rare events \cite{chekmarev04, trendelkamp14}. 
%
%
%
%
In contrast, constructing discrete-time Markov models from MD simulations to describe the slow, conformational dynamics of simple molecules has proven a successful strategy in many scenarios \cite{husic2018markov}.
%
%
Yet, discrete-time models of inherently continuous-time dynamical systems 
necessarily introduce a finite time scale, the so-called lag time (i.e.~the amount of time passing in between each step), 
%
which is an extra hyperparameter that has to be chosen by hand, and validated via secondary procedures.
Such an arbitrary discretization of time is rarely interpretable \cite{mcgibbon2015efficient}, and the corresponding lag time has to be sufficiently long to make the process Markovian, which lowers the resolution of the model \cite{koehs22}.

A natural alternative is to assume that the conformational dynamics of ADP is described by a hidden (continuos-time) MJP. 
%
In this section we take this route and train NeuralMJP on an all-atom MD simulation of ADP. This simulation dataset was originally used by \citet{wang2019machine} and was made available to us via private communication (see the acknowledgements below). 
%
%
Details about data preprocessing and training procedures are provided in Appendix~\ref{appendix:datadescription:adp}.

Let us choose a six-state NeuralMJP to model the ADP conformational dynamics. 
We compare our findings against those of three models:
(i) the VAMPnets model of \citet{mardt17}, 
%
(ii) the GMVAE model of \citet{varolguenes19}, 
and (iii) the estimates provided by the trajectory analysis of \citet{trendelkamp14}.

\textbf{Results}. NeuralMJP automatically splits the Ramachandran plot into the six regions shown in Figure \ref{fig:adp_6_states}. 
These states can be identified with known metastable ADP conformations, and are in good agreement with the states found by both VAMPnets and GMVAE \cite{mardt17, varolguenes19}.
%

From Table \ref{tab:adp_stationary} we find that the asymptotic dynamics inferred by our model, as measured by the stationary distribution, is comparable to the stationary distribution of the hidden Markov chain inferred by VAMPnets.
Much of the probability mass is concentrated on states $\rom{1}$ and $\rom{2}$, which correspond to low-energy regions of the Ramachandran plot. 
Also note that the stationary probabilities for states $\rom{5}$ and $\rom{6}$ are considerably lower than those for the other states. 
Transitions to these states represent the rare events of the ADP dynamics \cite{trendelkamp14}. 
Table \ref{tab:adp_time scales} reports the relaxation time scales characteristic of the ADP as inferred by all models.
Ours identifies time scales of three different orders of magnitude. The fast time scales inferred by us agree well with the fast time scales found by the baselines. 
The longest relaxation time is however short, as compared to what is extracted by the other models. 

A similar pattern emerges with our mean first-passage time estimates. These are gathered in Table \ref{tab:adp_mfps}.
Our model predicts more frequent transitions to the rare states $\rom{5}$ and $\rom{6}$, i.e. it predicts faster first-passage times, as compared to e.g.~the $60$ ns estimate reported in \citet{trendelkamp14}. 
And yet for the more recurrent $\rom{1} \Leftrightarrow \rom{4}$ transitions the estimates of both approaches are quite similar. 
These observations seem to indicate that NeuralMJP is biased towards fitting faster time scales (see also Section~\ref{sec:iondata} above for similar findings). 

All in all, our results show that NeuralMJP can correctly describe the conformational dynamics of simple molecules, without the need to introduce any artificial lag time scale.
We report additional results (as e.g.~experiments with a different number of hidden states, error bar estimations and further analysis of the model) in Appendix~\ref{appendix:datadescription:adp}.

%



\begin{table}[]
\caption{Relaxation time scales for six-state Markov models of ADP. The time scales are ordered by size and reported in nanoseconds. VAMPnet results are taken from \citet{mardt17}, GMVAE from \citet{varolguenes19} and MSM from \citet{trendelkamp14}.}
\label{tab:adp_time scales}
\vskip 0.15in
\begin{center}
\begin{small}
\begin{sc}
    \begin{tabular}{rccccc}
    \toprule
      & \multicolumn{5}{c}{Relaxation time scales (in $ns$)} \\
    \midrule
     VAMPnets  & $0.008$ & $0.009$ & $0.055$ & $0.065$ & $1.920$ \\
     GMVAE & $0.003$ & $0.003$ & $0.033$ & $0.065$ & $1.430$ \\
         MSM &    -        & -     & -         &      -&  $1.490$ \\
    \midrule
     NeuralMJP &  $0.009$ &  $0.009$ & $0.043$ &    $0.069$ &  $0.774$ \\
    \bottomrule
\end{tabular}
\end{sc}
\end{small}
\end{center}
\vskip -0.1in
\end{table}

\begin{table}[]
\caption{Mean first-passage times for six-state Markov models of ADP. The row and column labels are the states identified in Figure \ref{fig:adp_6_states}. The entry of row $i$ and column $j$ represents the mean first-passage time for transitions $i \rightarrow j$ reported in nanoseconds.}
\label{tab:adp_mfps}
\vskip 0.15in
\begin{center}
\begin{small}
\begin{sc}
\begin{tabular}{rrrrrrr}
\toprule
    $\tau_{ij}$ & $\rom{1}$ & $\rom{2}$ & $\rom{3}$ & $\rom{4}$ & $\rom{5}$ & $\rom{6}$ \\
\midrule
    $\rom{1}$ & $0.   $   & $0.028$     & $0.290$     & $0.107$     & $13.727$    & $15.864$\\
    $\rom{2}$ & $0.032$   & $0.   $     & $0.287$     & $0.103$     & $13.728$    & $15.866$\\
    $\rom{3}$ & $0.132$   & $0.134$     & $0.   $     & $0.063$     & $13.729$    & $15.859$\\
    $\rom{4}$ & $0.073$   & $0.074$     & $0.187$     & $0.   $     & $13.706$    & $15.844$\\
    $\rom{5}$ & $0.917$   & $0.913$     & $1.055$     & $0.932$     & $0.    $    & $3.959 $\\
    $\rom{6}$ & $0.804$   & $0.800$     & $0.952$     & $0.824$     & $2.563 $    & $0.    $\\
\bottomrule
\end{tabular}
\end{sc}
\end{small}
\end{center}
\vskip -0.1in
\end{table}

\section{Conclusions}
\label{sec:conclusions}

In this work we introduced a novel, neural-based variational inference algorithm for MJP. The model leverages NVI and NeuralODEs to encode, in an end-to-end fashion, noisy time series data into the parameters defining the master equation of the posterior MJP.
We empirically demonstrated that the model was able to successfully infer hidden MJP from synthetic, experimental and MD simulation data.
%

%

\section{Limitations}
\label{sec:limitations}

The main limitation we find with NeuralMJP is its bias towards better fitting the faster time scales of the empirical data.
We observed evidence of these when modelling both, the switching ion channel data and the alanine dipeptide MD simulation.
We speculate that this bias might arise from the KL evaluation, which favors larger transition rates. 
We leave an analysis of this feature for future work.

\section{Acknowledgement}

This research has been funded by the Federal Ministry of Education and Research of Germany and the state of North-Rhine Westphalia as part of the Lamarr-Institute for Machine Learning and Artificial Intelligence (LAMARR22B). 
We would like to thank Lukas K\"ohs for sharing the experimental ion channel data with us.
The actual experiment was carried out by Kerri Kukovetz and Oliver Rauh, while working in the lab of Gerhard Thiel of TU Darmstadt.
%
%
Similarly, we would like to thank Nick Charron and Cecilia Clementi, from the Theoretical and Computational Biophysics group of the Freie Universit\"at Berlin, for sharing the all-atom alanine dipeptide simulation data with us.
The simulation was carried out by Christoph Wehmeyer while working in the research group of Frank No\'e of the Freie Universit\"at Berlin.


\bibliography{NeuralMJP_icml2023_rebiber}
\bibliographystyle{icml2023}

\newpage
\appendix
\onecolumn

\section{Derivation of Master Equation}
\label{sec:master_eq_derivation}
We roughly follow \citet{gardiner09} for the derivation of the master equation.

Given some stochastic process $\{Z(t): t \in \mathbb{R}^+ \}$ taking values on some countable (and possibly unbounded) set $\mathcal{Z}$, the Markov property states that the conditional probability of observing any future state of the process depends entirely on its current state. 

Thus the probability of observing $Z(t)=x$, given that $Z(t') = y$ and $Z(t'') = z$, with $t \ge t' \ge t''$, is given by
\begin{equation*}
    p(  x, t| y, t'; z, t'') = p(x, t| y, t').
\end{equation*}

The \textbf{instantaneous probability rate} of transitioning from state $z^\prime \in \mathcal{Z}$ to any other state $z \neq z^\prime \in \mathcal{Z}$ at $t \in \mathbb{R}^+$ is defined as 
\begin{equation}
     f(z \mid z^\prime, t) = \lim_{\Delta t \rightarrow 0}  \frac{1}{\Delta t} p(z, t+\Delta t| z', t).
     \label{eq:process_rate_app}
\end{equation}

The \textbf{forward master equation} describes how the conditional probabilities of $Z$ evolve with time. 
This evolution is completely described by the instantaneous probability rates:

\begin{align}
\nonumber
    \partial_t p(x, t)  & = \lim_{\Delta t \rightarrow 0}  \frac{1}{\Delta t} \left[ p(x, t+\Delta t) - p(x, t)  \right] \\ \nonumber
    & = \lim_{\Delta t \rightarrow 0}  \frac{1}{\Delta t} \left[ \sum_{z} p(x, t+\Delta t | z, t) p(z,t) - p(x, t)  \right] \\ \nonumber
    & = \lim_{\Delta t \rightarrow 0}  \frac{1}{\Delta t} \left[ \sum_{z \neq x} p(x, t+\Delta t | z, t) p(z,t) - \left( 1- p(x, t+\Delta t | x, t) \right) p(x, t)  \right] \\ \nonumber
    & = \lim_{\Delta t \rightarrow 0}  \frac{1}{\Delta t} \left[ \sum_{z \neq x} p(x, t+\Delta t | z, t) p(z, t) - \left(\sum_{z \neq x} p(z, t+\Delta t| x, t) \right)p(x, t)  \right] \\
    & = \sum_{z \neq x} \left[f(x|z, t) p(z, t) -  f(z|x, t) p(x, t)\right]
    \label{eq:forward}
\end{align}

Note that we used the law of total probability in the second and fourth line above.

Moreover, if $\mathcal{Z}$ is finite, the master equation \ref{eq:forward} can be written in matrix form.
Let $\*p(t) = \left( p(x, t) \right)_{x \in \mathcal{Z}}$ be the vector containing the time-dependent probabilities over the states in $\mathcal{Z}$ and $\dot{\*p}(t)$ its time derivative.
Then we can define the \textbf{rate matrix} or $\mathbf{\mathcal{Q}}$\textbf{-matrix} $\*F(t) \in \mathbb{R}^{\mid \mathcal{Z} \mid \times \mid \mathcal{Z} \mid}$ by
\begin{equation*}
    \*F_{zx}(t) = \begin{cases}
            -\sum_{z^\prime \neq x} f(z^\prime|x, t) & \text{if} \, \, x = z \\
            f(x| z, t), & \text{otherwise}.
    \end{cases}
\end{equation*}

and express the forward master equation by
\begin{equation*}
    \dot{\*p}(t) = \*p (t) \cdot \*F(t) .
\end{equation*}

\section{On Stationary Distribution, Relaxation Times and Mean First-Passage Times of MJPs}
\label{sec:stationarity-mfpt}
A MJP with \textit{time-independent} transition rates is called \textbf{homogeneous}. 
Let $\mathcal{Z}$ be the finite state space of a homogeneous MJP $Z$ with transition rates $f$ and let $\*F$ be the associated (time-independent) rate matrix.

\textbf{Stationary Distribution:}
A probability distribution $\*p^*$ on $\mathcal{Z}$ is called a \textbf{stationary distribution} of $Z$, if $\*p^* \cdot \*F = \*0$. 
Put differently, a stationary distribution is a left eigenvector of $\*F$ with eigenvalue $0$, that is also a probability distribution. 
If $Z$ is \textbf{irreducible}, roughly meaning all states are connected, there exists a unique stationary distribution $\*p^*$. 
Moreover, $Z$ will converge to $\*p^*$ from any initial distribution.
See \cite{tolver16} for a detailed derivation of these results.

We find that the learned prior MJPs in our experiments are irreducible and will therefore converge to their stationary distribution in the long-term. 
However, short-term predictions with the prior MJP still significantly depend on the observed time series. 

\textbf{Relaxation Time:} 
Let $\lambda_2, \lambda_3, \dots, \lambda_{\left|\mathcal{Z}\right|}$ be the non-zero eigenvalues of $\*F$. 
The \textbf{time scales} of the MJP are defined as $\left|{\operatorname{Re}\lambda_2}\right|^{-1}, \left|{\operatorname{Re}\lambda_3}\right|^{-1}, \dots, \left|{\operatorname{Re}\lambda_{\left|\mathcal{Z}\right|}}\right|^{-1}$. 
They are exponential rates of decay that determine convergence from any state of the process to its stationary distribution.  

The long-term convergence behaviour is dominated by the largest time scale, which is known as the \textbf{relaxation time} of the process.

\textbf{Mean First-Passage Times:} 
Suppose $Z(0) = i \in \mathcal{Z}$, then the first-passage time to state $j \in \mathcal{Z}$ is defined as 
\begin{equation*}
    T_{ij} = \inf \{ t \geq 0: Z(t) = j \mid Z(0) = i \}
\end{equation*}
and the associated \textbf{mean first-passage time} as 
\begin{equation*}
  \tau_{ij} =\mathbb{E}[T_j \mid Z(0) = i] \quad .  
\end{equation*}

One can show that for finite state, time-homogeneous MJP, mean first-passage times are the solutions of $\left| \mathcal{Z} \right|$ systems of equations:
\begin{equation*}
    \begin{cases}
        \tau_{ii} = 0 & \\
        1 + \sum_{k} \*F_{ik} \tau_{kj} = 0  &, j \neq i 
    \end{cases}
\end{equation*}

See Section $4.4$ of \cite{durrett99} for a derivation of this result.

\begin{figure*}[b!]
    \centering
    \pdftooltip{\includegraphics[scale=0.9]{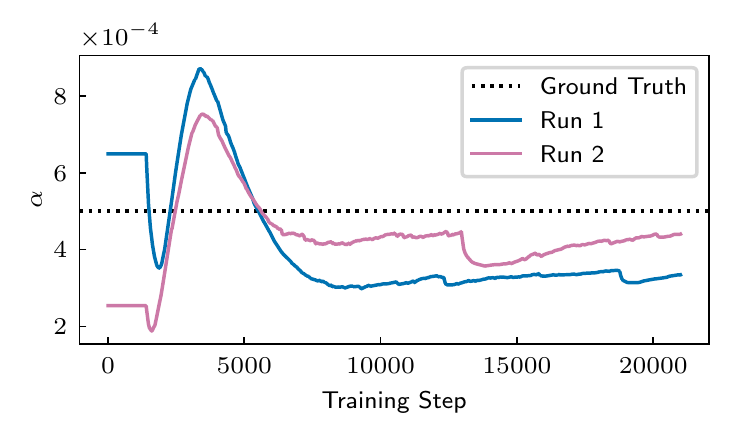}}{
    Evolution of trainable Lotka-Volterra parameter $\alpha$ for two runs with different initializations. Roughly $20000$ training steps are recorded and $\alpha$ varies between $0.0002$ and $0.0009$ during training. The initial difference of $\alpha$ in the two runs is $0.0004$ and remains at about $0.0001$ after $8000$ steps till the end of training.
    }
    \caption{Evolution of the Lotka-Volterra trainable parameter $\alpha$ during training of NeuralMJP for two random initializations. 
Given that the prior parameters are learned separately from the rest (see Algorithm~\ref{alg:training}), $\alpha$ is learned only later in training and thus constant at the beginning. 
The gap in $\alpha$ of the two runs is not consistently closed and remains significant throughout training. }
    \label{fig:implicit_explicit_prior}
\end{figure*}

\section{Implicit vs Explicit Parametrization of The Prior Transition Rates}\label{app:implicit_prior}
During early development of NeuralMJP, we treated the entries of the prior transition matrix as trainable parameters. 
One of the main empirical observations we made was that, during training, their values changed very slowly, and eventually got stuck in different regions of parameter space. 
We provide an example of this behaviour in Figure \ref{fig:implicit_explicit_prior}, for the $\alpha$ parameters of the Lotka-Volterra process (see Appendix~\ref{appendix:datadescription:lv} for details).

Key here is that the location of these regions was strongly dependent on the parameters' initialization values.
Assuming no prior knowledge about the scale of the entries of the (prior) transition matrix, as is the case in both the experimental ion channel data and the molecular dynamics simulation, one cannot but choose the initialization values randomly, which led to inconsistencies in the inferred rates.

We speculate that the reason behind the slow dynamics of the prior parameters was the magnitude of the gradient signals, which directly modified their values. 
Opposite to the simple trainable parameters, a generator network can leverage weak gradient signals to produce large changes in its output values, as to better fit the posterior rates. 
We found in practice that this was indeed the case.

\section{Derivation of Variational Bound}
\label{sec:elbo-derivation}
In this section we derive Eq.~\ref{eq:loss}, a variational lower bound on the logarithm of the marginal likelihood of the data and the objective of NeuralMJP. 
This bound was already used by \citet{opper07} for their expectation-maximization approach. 
See \cite{cohn09master} for a similar derivation of this bound. 

We drop the dependence on parameters $\psi$, $\phi$ from the subscripts and the dependency of posterior transition rates $g$ on the observations for a less cluttered derivation.

Consider observations $\*x_0, \dots, \*x_N$ recorded at observation times $0 \leq \tilde{t}_0 < \dots < \tilde{t}_N \leq T$. 
Let $t_{0:K}$ be a discretization $0 = t_0 < t_1 < \dots < t_K = T$ of the interval $[0,T]$ such that $\Delta t_k = t_{k+1} - t_k$ is small and bounded from above by some $\overline{\Delta t} > 0$. 
Moreover, assume that, without loss of generality, observation times $\tilde{t}_i$ are part of the discretization, so there exists a map $m: \{0, \dots, N \} \longrightarrow \{0,\dots,K\}$ such that $\tilde{t}_i = t_{m(i)}$. 
Let $z_{0:K}$ denote a trajectory of latent states recorded at the discretization grid and $\sum_{z_{0:K}}$ the sum over all such trajectories.

By the assumptions on the generative model and Jensen's inequality, we have
\begin{align}
\label{eq:objective_jensen}
    \log p(\*x_0, \*x_1, \dots, \*x_N) &= \log \left( \sum_{z_{0:K}} \frac{q(z_{0:K}, t_{0:K})p(z_{0:K}, t_{0:K})p(\*x_0, \*x_1, \dots, \*x_N \mid z_{0:K}, t_{0:K})}{q(z_{0:K}, t_{0:K})}\right) \nonumber \\
                &\geq \sum_{i=0}^N \mathbb{E}_{q(z,t_{m(i)})} [\log p(\*x_i \mid z_{m(i)})] - \text{KL}_\text{discr}[q(z_{0:K}, t_{0:K}) \Vert p(z_{0:K} , t_{0:K})]
\end{align}
with 
\begin{align*}
    \text{KL}_\text{discr}[q(z_{0:K}, t_{0:K}) \Vert p(z_{0:K}, t_{0:K})] = \sum_{z_{0:K}} q(z_{0:K}, t_{0:K}) \log \frac{q(z_{0:K}, t_{0:K})}{p(z_{0:K}, t_{0:K})} \quad .
\end{align*}

Note that the first summand of Eq.~\ref{eq:objective_jensen}, referred to as the reconstruction cost, already appears in Eq.~\ref{eq:loss}, using the notation $z(\tilde{t}_i) = z_{m(i)}$. 

We now focus on the second summand. 
Because of the Markov property, we can factorize prior and posterior probabilities in $\text{KL}_\text{discr}[q(z_{0:K}, t_{0:K}) \Vert p(z_{0:K}, t_{0:K})] $. 
After rearranging some terms, each summand is independent of future trajectories, simplifying the expression. 
\begin{align}
\label{eq:objective_factorized}
     &\text{KL}_\text{discr}[q(z_{0:K}, t_{0:K}) \Vert p(z_{0:K}, t_{0:K})]   \nonumber \\
     =&\sum_{z_{0:K}} q(z_0, t_0) q(z_1, t_1 \mid z_0, t_0) \dots q(z_K, t_K \mid z_{K-1}, t_{K-1}) \log \frac{q(z_0, t_0) q(z_1, t_1 \mid z_0, t_0) \dots q(z_K, t_K\mid z_{K-1}, t_{K-1}) }{p(z_0, t_0) p(z_1, t_1 \mid z_0, t_0) \dots p(z_K, t_K\mid z_{K-1}, t_{K-1}) } \nonumber \\
     =&\sum_{z_0} q(z_0)\log \frac{p(z_0)}{q(z_0)} + \sum_{k=0}^{K-1} \sum_{z_k} q(z_k , t_k) \sum_{z_{k+1}} q(z_{k+1}, t_{k+1} \mid z_k, t_k) \log \frac{q(z_{k+1}, t_{k+1} \mid z_k, t_k)}{p(z_{k+1}, t_{k+1} \mid z_k, t_k)} \nonumber\\
     =&\sum_{z_0} q(z_0)\log \frac{p(z_0)}{q(z_0)} + \sum_{k=0}^{K-1} \Delta t_k \sum_{z}\sum_{z^\prime} \frac{q(z , t_k)  q(z^\prime, t_{k+1} \mid z, t_k)}{\Delta t_k} \log \frac{q(z^\prime, t_{k+1} \mid z, t_k)}{p(z^\prime, t_{k+1} \mid z, t_k)}
\end{align}
In the last equation, we changed the notation in preparation of the limit $\Delta t_k \rightarrow 0$ and the Riemann integral below. 

We consider the summands of Eq.~\ref{eq:objective_factorized} individually. 

Following \citet{opper07}, our model does not include a prior for the initial condition, hence the first summand vanishes. 

For the other summand, we first consider the case of $z^\prime \neq z$. 
Recall that by Eq.~\ref{eq:process_rate_app} and the time-homogeneous prior, we have
\begin{align*}
    q(z^\prime, t_{k+1} \mid z, t_k) &= g(z^\prime \mid z, t_k) \Delta t_k + o(\Delta t_k) \\
    p(z^\prime, t_{k+1} \mid z, t_k) &= f(z^\prime \mid z) \Delta t_k + o(\Delta t_k)
\end{align*}
for small $\Delta t_k$. %
Therefore: 
\begin{align}
\label{eq:case_zprime_eq_z}
   &  \frac{q(z , t_k)  q(z^\prime, t_{k+1} \mid z, t_k)}{\Delta t_k} \log \frac{q(z^\prime, t_{k+1} \mid z, t_k)}{p(z^\prime, t_{k+1} \mid z, t_k)}  \nonumber \\
   =& q(z, t_k) \left[ g(z^\prime \mid z, t_k) + \frac{o(\Delta t_k)}{\Delta t_k}\right] \log \frac{g(z^\prime \mid z, t_k) + \frac{o(\Delta t_k)}{\Delta t_k}}{f(z^\prime \mid z) + \frac{o(\Delta t_k)}{\Delta t_k}} \nonumber\\
   &\overset{\Delta t_k \rightarrow 0 }{\longrightarrow} q(z, t_k) g(z^\prime \mid z, t_k) \log \frac{g(z^\prime \mid z, t_k)}{f(z^\prime \mid z)}
\end{align}

We now consider the case of $z^\prime = z$, where we use the approximation $\log(1+y) = y + o(y)$ for small $y$. %
By the law of total probability we can express $q(z, t_{k+1} \mid z, t_k)$ as
\begin{align*}
    q(z, t_{k+1} \mid z, t_k) &= 1 - \sum_{z^\prime \neq z} q(z^\prime, t_{k+1} \mid z, t_k) \\ 
        & = 1 - \Delta t_k \sum_{z^\prime \neq z} g(z^\prime \mid z, t_k) - o(\Delta t_k) 
\end{align*}

and similarly $p(z , t_{k+1} \mid z, t_k)$. 
Therefore:
\begin{align}
\label{eq:case_z_eq_z}
    &  \frac{q(z , t_k)  q(z, t_{k+1} \mid z, t_k)}{\Delta t_k} \log \frac{q(z, t_{k+1} \mid z, t_k)}{p(z, t_{k+1} \mid z, t_k)} \nonumber \\
   =& q(z, t_k) \left[1 - \Delta t_k \sum_{z^\prime \neq z} g(z^\prime \mid z, t_k) - o(\Delta t_k) \right] \left[ - \sum_{z^\prime \neq z} g(z^\prime \mid z, t) + \sum_{z^\prime \neq z } f(z^\prime \mid z) + \frac{o(\Delta t_k)}{\Delta t_k}\right] \nonumber\\
   &\overset{\Delta t_k \rightarrow 0}{\longrightarrow} q(z, t_k) \left[ \sum_{z^\prime \neq z} f(z^\prime \mid z) - g(z^\prime \mid z, t_k)\right]
\end{align}

Finally, taking the limit $\overline{\Delta t} \rightarrow 0 $ and combining Eq.~\ref{eq:objective_jensen} with \ref{eq:case_zprime_eq_z} and \ref{eq:case_z_eq_z} yields the variational bound
\begin{align*}
    &\log p(\*x_0, \*x_1, \dots, \*x_N) \geq \\
    &\sum_{i=0}^N \mathbb{E}_{q(z,\tilde{t}_i)} [\log p(\*x_i \mid z(\tilde{t}_i))]  - \int_0^T dt \mathbb{E}_{q(z,t)} \sum_{z^\prime \neq z} \left\{ f(z^\prime \mid z) - g(z^\prime \mid z, t)  + g(z^\prime \mid z, t) \log \frac{g(z^\prime \mid z, t)}{f(z^\prime \mid z)} \right\} .
\end{align*} 

\section{Two-Step Optimization Scheme}
\label{app:2-step-prior}
We describe the training scheme for our implementation of NeuralMJP in Algorithm \ref{alg:training}. 
Notably, we employ a two-step update approach that trains the encoder-decoder pair and prior parameters separately. 

In principle, one can train the encoder-decoder pair together with the prior parameters in a single back-propagation step. 
However, we have found empirically that the KL regularization had a smoothing effect on the inferred, time-dependent (i.e. inhomogeneous) posterior rates. 
This effect generally led the model to underestimate the time-independent (i.e. homogeneous) transition rates in the ground-truth experiments and, in some cases, it even led to mode collapse.

By construction, the posterior process is restricted to be an inhomogeneous MJP, even without the KL regularization.
Simultaneously updating the encoder-decoder pair while keeping the prior components fixed, and only updating the latter at a second step, allowed us to find the best inhomogeneous MJPs fitting the data. 
Thus the main purpose of the isolated KL term is to drive the prior towards homogeneous MJPs that best fit the optimal, inhomogeneous posterior process.

In practice we found this two-step update approach to yield excellent results with respect to the ground-truth processes, as compared to the baselines. 

\section{Training Instabilities And Index Collapse}
\label{app:index_collapse}
In experiments with gaussian emission models, NeuralMJP can exhibit behaviour comparable to index collapse \cite{bengio2015conditional, shazeer2017, kaiser18, schwab2019granger, ojeda2021switching}. 
The posterior MJP of such models never assigns any probability mass to some states. 

This defect affects the time series reconstructions of these models, which gets partially compensated by larger learned covariances in the emission model. 
However, its effect on the posterior rates extends to the prior rates, which can not be compensated. 

In our experiments, we chose the number of latent states intentionally, either to recover some ground-truth or to compare with the results of other papers. 
Therefore, we want to avoid index collapse in our experiments and use two schedulers to counteract this behaviour.

The first scheduler is directly connected to the gaussian emission models. 
At the start of training, we fix the covariance to a constant, diagonal covariance matrix and only learn the gaussian mean. 
Only after the model has arranged the latent states and their corresponding mean reconstruction, we start learning the covariance matrix.

The second scheduler is connected to the latent dynamics and is inspired by a remark from \citet{kidger21}. 
At the beginning of training, randomly initialized posterior rates do not represent the observed dynamics well. 
This can lead the model towards local minima, where the latent space is not suitably arranged. 

Therefore, we only train on the first $10$ observations of each time series for the first $3000$ batches. 
Afterwards, the number of observations is annealed to the full time series over a period of $5000$ batches. 
For the Alanine Dipeptide Protein dataset, we increased this period to $10000$ batches. 

In our experiments, the combination of these schedulers reduced the occurrence of index collapse noticeably.

\section{How to Sample Markov Jump Processes Trajectories?}
\label{app:sampling_homog_mjp}
The algorithm for sampling trajectories of homogeneous MJPs is due to \citet{gillespie77}. 
Here we recall a derivation of the algorithm for convenience. 

Let $Z$ be a homogeneous MJP with finite state space $\mathcal{Z}$ and time-independent transition rates $f$. 
Assume $Z$ is at state $y \in \mathcal{Z}$ at time $t$, so $Z(t)=y$, and $f(z \mid y) \neq 0$ for at least one $z \in \mathcal{Z}$. 

To sample a trajectory of $Z$ starting at $t$, it is enough to repeatedly sample a single transition, because of the Markov property. 
Moreover, to sample the transition of $Z$ away from $y$, it is enough to sample the arrival time of the next transition and the corresponding state transition. 

Let $T(y,t) \geq t$ be the time of the next transition of $Z$, given $Z(t) = y$. 
To find the distribution of $T(y,t)$, consider $Q(y, t, t + \tau) = \mathbb{P}[T(y,t) \geq t + \tau]$ for some $\tau \geq 0$. 

By construction of the transition rates $f$ (Eq.~\ref{eq:process_rate_app}), we have
\begin{equation*}
   p(z, \tau + \Delta t \mid y, \tau) = f(z \mid y) \Delta t + o(\Delta t)  
\end{equation*}
for $z \neq y$ and small $\Delta t$. 
Hence, the following transformations are valid: 
\begin{align*}
Q(y, t + \tau, t + \tau + \Delta t) &= \prod_{z \neq y} \mathbb{P}[\text{No transition from $y$ to $z$ in } [t+\tau,t+ \tau + \Delta t]] \\
&= \prod_{z \neq y} (1 - p(z, t+\tau + \Delta t \mid y, t+\tau)) \\
&= \prod_{z \neq y} (1 - f(z \mid y) \Delta t + o(\Delta t)) \\
&= 1 - \sum_{z \neq y} f(z \mid y) \Delta t + o(\Delta t)
\end{align*}

Therefore, 
\begin{align*}
    Q(y, t, t + \tau + \Delta t) &= Q(y, t, t +\tau) Q(y, \tau, t+ \tau + \Delta t) \\
    &= Q(y, t, \tau)(1 - \sum_{z \neq y} f(z \mid y) \Delta t + o(\Delta t))
\end{align*}
and after some rearranging and $\Delta t \rightarrow 0 $,
\begin{equation*}
    \partial_\tau Q(y,t,t + \tau) = - \sum_{z \neq y} f(z \mid y) Q(y, t, t + \tau)
\end{equation*}

With the assumptions from above we have $Q(y,t,t)=1$. 
Solving the initial value problem yields 
\begin{equation*}
    Q(y, t, t + \tau) = e^{-\sum_{z \neq y} f(z \mid y) \tau}
\end{equation*}
and hence
\begin{align*}
    \mathbb{P}[T(y,t) = \tau] &= \partial_\tau [1-\mathbb{P}[T(y,t) \geq \tau]] \\
    &= \sum_{z \neq y} f(z \mid y)  e^{-\sum_{z \neq y}f(z \mid y) \tau} 
\end{align*}

So $T(y,t)$ is exponentially distributed on values $[t, \infty )$ with rate $\lambda = \sum_{z \neq y} f(z \mid y)$ and can be sampled efficiently. 

Also following Eq.~\ref{eq:process_rate_app}, the probability of transitioning from $y$ to $x$ must be
\begin{equation}
    \frac{f(x \mid y)}{\sum_{z \neq y} f(z \mid y)} \quad .
    \label{eq:state_transition_prob}
\end{equation}

Thus, sampling one transition of a MJP at state $y$ at time $t$ requires sampling a exponentially distributed waiting time from $\text{Exp}(\lambda)$ and distribution \ref{eq:state_transition_prob} over $\mathcal{Z} \setminus \{y\}$. 

Note that this algorithm also applies to MJPs with countable state space $\mathcal{Z}$, if for every $z \in \mathcal{Z}$ only finitely many transition rates $f(y \mid z)$ are non-zero, like in the Lotka-Volterra process (see Appendix~\ref{appendix:datadescription:lv} for details).

Furthermore, this sampling algorithm implies that MJP trajectories are right-continuous and piecewise-constant.

\section{Reproducibility and Experiments}
\label{app:reproducibility_and_experiments}
\subsection{Dataloading}
Because of data generation and pre-processing, all time series in train, test and validation sets of a given dataset are of the same length. 
Each observation is comprised of a observation time and a observation value. 
We enrich them by the time difference to the subsequent observation. 
For the last observation of each time series, his information is missing, so we drop it during dataloading.

\subsection{Architectures and Hyperparameters}
The network architectures described below are kept constant across all experiments, apart from the LV single time series dataset, where we reduced the MLPs to a single hidden layer of size $64$. 

We fix the dimension $H$ of  $\*h_T$ to 256 in all experiments, apart from the LV single time series dataset, where it is set to $64$. 

MLPs of the ODE-RNN and the posterior process $\+\Psi_\theta$ consist of hidden layers $[256,256]$ and $[256,256,128]$ respectively, have Tanh activation functions and are initialized with zero bias and weights sampled from $\mathcal{N}(0, 0.01)$. 
The MLP $\+\Lambda_\phi$ has hidden layers $[128,128]$ with ReLU activation functions and is initialized with Kaiming initialization. 
If the emission model contains a MLP, it has hidden layers $[128,128]$, ReLU activation functions and is initialized with zero bias and weights sampled from $\mathcal{N}(0, 0.01)$.

The MLP for the generative prior $\Phi_\theta$ consist of one layer of width $64$ with ReLU activation function and is initialized with Kaiming initialization. 
Its input is a $64$-dimensional sample from $\mathcal{N}(0, 0.01)$. 

We use layer normalization and dropout of $0.2$ for all MLPs. 

\subsection{NeuralODEs: Additional Details}

We use the torchdiffeq python package \cite{torchdiffeq} to solve all NeuralODEs in our model. 
The ODE-RNN is solved with the Runge-Kutta method. 
The posterior master equation is solved with the adaptive step Dormand-Prince method and tolerances of $0.001$.
For faster training on the ADP dataset, we set the tolerances to $0.01$ for these experiments. 

To handle batches of irregular time series, we apply the technique described in Appendix F of \cite{chen20}. 
 
We use STEER from \cite{gosh20} to regularize the ODE-RNN. 

\subsection{On The Variational Posterior Process}

The posterior transition rates $\*g_{\phi}$ are time-dependent. 
We use Mercer a embedding \cite{xu19} of $t$ with $10$ frequencies and $20$ Fourier coefficients as the time component of the input of $\+\Psi_{\phi}$, the MLP for $\*g_{\phi}$. 

Note that the Mercer embedding is not an integral part of our model. 
We found anecdotally that their addition leads to marginal improvements in the learned dynamics, but do not change the results fundamentally.

The marginal posterior distribution at observation times are sampled with the Gumbel-Softmax estimator from \cite{jang16} with a sampling temperature of $1$.

\subsection{Numerical Approximation of the Kullback-Leibler Divergence}
\label{appendix:sampling_prior}
We approximate the integration of the Kullback-Leibler divergence in Eq.~\ref{eq:loss} by Gaussian quadrature with $200$ points.

Alternatively, it could be approximated by augmenting the dynamics of the posterior master equation with the integrand of the Kullback-Leibler divergence. 
In practice, we found this method to be much slower without providing noticeable benefits. 

Finally, we tried approximating the KL with Tanh-Sinh quadrature. 
Its main advantage over Gaussian quadrature is the progressive use of integrand evaluations when lowering the stepsize. 
We can not take advantage of this property, because any additional integrand evaluation requires solving the posterior master equation, which is computationally expensive.

\subsection{Optimizers}
We use the Adam optimizer and apply gradient clipping with global norm $1$ for all experiments. 
However, learning rate and possible learning rate annealing depend on the specific experiments.

For the single LV time series and the Two-Mode Hybrid System we use a learning rate of $0.001$. 

For LV, DFR and ADP data we use a learning rate of $0.001$, annealed by a factor of $0.8$ every $50$ epochs. 

For Simple Protein Folding Model and Switching Ion Channel Data we use a learning rate of $0.0001$, with an annealing of $0.8$ every $100$ and $200$ epochs respectively. 

In experiments with the Switching Ion Channel Data we found that the generative prior does not converge fast enough with this annealing schedule. 
Hence, we use a separate optimizer only for the corresponding MLP with constant learning rate $0.0005$.

\subsection{Metrics}
To compare RMSE between test and prediction sets, we use the following RMSE for multivariate time series. 

Assume there are $N$ time series of length $T$ in $D$ dimensional space. 
Let $x_{ijk}$ be the observed $k$-th dimension of observation $j$ of the $i$-th time series. 
Let $\hat{x}_{ijk}$ be the prediction of $x_{ijk}$. 
Then we define:
\begin{equation*}
    RMSE = \sqrt{\frac{1}{NT} \sum_{i=1}^N \sum_{j=1}^T \sum_{k=1}^D (\hat{x}_{ijk} - x_{ijk})^2}
\end{equation*}

Next, we consider the RMSE of the prediction at a specific step $j$ of all $N$ time series:
\begin{equation*}
     RMSE_j = \sqrt{\frac{1}{N} \sum_{i=1}^N \sum_{k=1}^D (\hat{x}_{ijk} - x_{ijk})^2}
\end{equation*} 

Note that this is the above formula with $T=0$. \par

To evaluate the capabilities on the first $m$ steps we compute the mean RMSE of the first $m$ predictions:
\begin{equation*}
    \frac{1}{m} \sum_{j=1}^m RMSE_j
\end{equation*}

\section{Training times and resource consumption}
\label{app:training_times}
We summarize the training times for all experiments in Table \ref{tab:app-training-times}. 
Some models only only after several days of training, despite loose tolerances. 
Presumably, more modern NeuralODE implementations like diffrax \cite{kidger21} speed up training of our model significantly. 
However, note that these training times are on the same scale as those of other NeuralODE based models.

Each experiment was performed on a single NVIDIA GeForce GTX 1080 Ti.

\subsection{Computational Complexity}
The structure of NeuralMJP, a combination of a backward in time ODE-RNN encoding and a forward in time NeuralODE solve, is the structure of LatentODE \cite{rubanova19}. 
Thus, the computational complexity of NeuralMJP and LatentODE is equivalent. 
Like the ODE-Net \cite{chen18}, it scales linearly with the number of function evaluations of the ODE solvers in the forward pass of the model. 

In our implementation of NeuralMJP, the master equation is solved with an adaptive step solver. 
The number of steps it takes depends on the complexity of the underlying dynamics.

\begin{table}[h]
\caption{Training times for selected runs of all NeuralMJP experiments, including the number of datapoints in their training data.}
\label{tab:app-training-times}
\vskip 0.15in
\begin{center}
\begin{small}
\begin{sc}
\begin{tabular}{cccc}
\toprule
        Data     &    Experiment &   Number of datapoints $\times 10^3$ &  Training time (hours) \\
\midrule
\multirow{4}{*}{DFR}	    & Irregular grid	& 224		& 41.3 \\
	    & Shared grid		& 224		& 22.5 \\
	    & Regular grid		& 224		& 31.2 \\
        & Single Time Series& 0.016          & 36.61 \\
\midrule
\multirow{2}{*}{Ion Channel} & One-Second Window		& 5		    & 21.5 \\
	    & Full Dataset		& 162.6		& 22.0 \\
\midrule
\multirow{3}{*}{LV}		& Irregular grid	& 224		& 33.1 \\
	    & Shared grid		& 224		& 19.5 \\
	    & Regular grid		& 224		& 19.0 \\
\midrule
\multirow{3}{*}{ADP}		& Six States          & 896		& 48.1 \\
        & Three States          & 896       & 20.6 \\
        & Two States          & 896       & 19.4 \\
\midrule
Simple Protein Folding Model &         & 89.6       & 26.8 \\
\midrule
Two-Mode Hybrid System &    & 0.067     & 26.3 \\
\bottomrule
\end{tabular}
\end{sc}
\end{small}
\end{center}
\vskip -0.1in
\end{table}

\newpage

\section{Discrete Flashing Ratchet}
\label{app:dfr}
\begin{table}[b!]
\caption{DFR parameters learned by NeuralMJP on multiple datasets with different observation grids. Additionally, KL divergence and reconstruction loss on the test set. Last row contains the results on \textsc{Irregular Grid Data} with constrained posterior process. Values are reported with mean and standard deviation of $5$ runs of NeuralMJP with different initializations.}
\label{tab:app-additonal-dfr}
\vskip 0.1in
\begin{center}
\begin{small}
\begin{sc}
\begin{tabular}{rccccc}
\toprule
                 &                 KL &               NLL &               $V$ &               $r$ &               $b$ \\
\midrule
          Ground Truth &                  - &                 - &           $1.00$ &           $1.00$ &           $1.00$ \\
\midrule
   Irregular Grid Data & $20.0 \pm 1.1$ & $0.21 \pm 0.01$ & $0.98 \pm 0.05$ & $1.11 \pm 0.06$ & $1.13 \pm 0.04$ \\
      Shared Grid Data & $18.7 \pm 0.5$ & $0.14 \pm 0.01$ & $0.95 \pm 0.03$ & $1.18 \pm 0.04$ & $1.14 \pm 0.06$ \\
     Regular Grid Data & $21.8 \pm 0.7$ & $0.18 \pm 0.02$ & $1.00 \pm 0.05$ & $1.37 \pm 0.05$ & $1.36 \pm 0.05$ \\
Masked Posterior Rates & $17.1 \pm 0.4$ & $0.22 \pm 0.01$ & $0.99 \pm 0.02$ & $1.18 \pm 0.04$ & $1.16 \pm 0.03$ \\
\bottomrule
\end{tabular}
\end{sc}
\end{small}
\end{center}
\end{table}
\subsection{Data Description and Pre-Processing}
The Discrete Flashing Ratchet (DFR) process \cite{roldan10} is a MJP that models the position of a particle in a termal bath of temperature $T$ subject to a periodic, asymmetric potential.
The particle can attain states $0, 1$ or $2$. 
The potential can be \textsc{ON} or \textsc{OFF}, yielding six possible states for the MJP.
The transition rates of the DFR process are \cite{roldan10}
\begin{eqnarray}
f(j, \text{ON} \mid i, \text{ON}) & = & \exp\left(- \frac{\beta V}{2}(j - i)\right) \quad \text{for} \,\, \,  i \neq j, \nonumber \\
f(j, \text{OFF} \mid i, \text{OFF}) &= &b, \quad \text{for} \,\, \, i \neq j, \nonumber \\
f(i, \text{OFF} \mid i, \text{ON}) &= &f(i, \text{ON} \mid i, \text{OFF})= r, \nonumber 
\end{eqnarray}
where $\beta = \frac{1}{T}$ and $V$, $r$ and $b$ are parameters of the model. 

We sample trajectories of DFR with parameters $T=1$, $V=1$, $r=1$ and $b=1$ and a time window of $[0, 2.5]$. 
Initial states are sampled from the stationary distribution of the process: 
\begin{align*}
p(0,\text{\textsc{On}})&=0.3012    &   p(1,\text{\textsc{On}})&=0.1365    &   p(2,\text{\textsc{On}})&=0.0623 \\
p(0,\text{\textsc{Off}})&=0.2003    &   p(1,\text{\textsc{Off}})&=0.1591   &   p(2,\text{\textsc{Off}})&=0.1406
\end{align*}

We sample $5000$ trajectories of DFR with the algorithm described in Appendix \ref{app:sampling_homog_mjp} and observe them at three observation time grids of length $50$. 
For the \textsc{Irregular Grid} dataset, the observation times for every time series are sampled uniformly in a given time window. 
For the \textsc{Shared Grid} dataset, observation times are sampled uniformly in a given time window, and then used as a time series for all trajectories. 
Finally, for the \textsc{Regular Grid} dataset, all time series have a fixed, regular grid with equidistant observation times.
We do not add noise to the observed states, treat them as categorical observations and load them as one-hot vectors.

Out of the $5000$ time series, we use $70$ batches of size $64$ for training and leave the remaining time series for test and validation. 
During training, we normalize observation times to lie in $[0,1]$.

We simulate $100$ additional trajectories on a time window twice as long as above for prediction. 
The first half of each trajectory is treated as another test set and observed on the same grid as the training data. 
The second half is also observed at $50$ observation times on the same type of grid. 
Thus, each time series in the prediction set is of length $100$.

\subsection{Modelling and Results}
We trained NeuralMJP with a six-state, unconstrained posterior MJP, meaning transitions between all states are possible. 
The prior is constrained to a DFR process and parametrized as above. 
On the \textsc{Irregular Grid} dataset, we additionally trained a NeuralMJP, in which the posterior MJP is constrained to transitions which are possible in the prior process.

In the absence of noise, we train these NeuralMJPs without an emission model and use cross-entropy as reconstruction loss. 
We summarize our results in Table \ref{tab:app-additonal-dfr}. 
Constraining the posterior MJP reduces the KL loss, but the inferred parameters are slightly more inaccurate.

\newpage

\begin{figure}[h]
    \centering
    \pdftooltip{\includegraphics[width=0.7\textwidth]{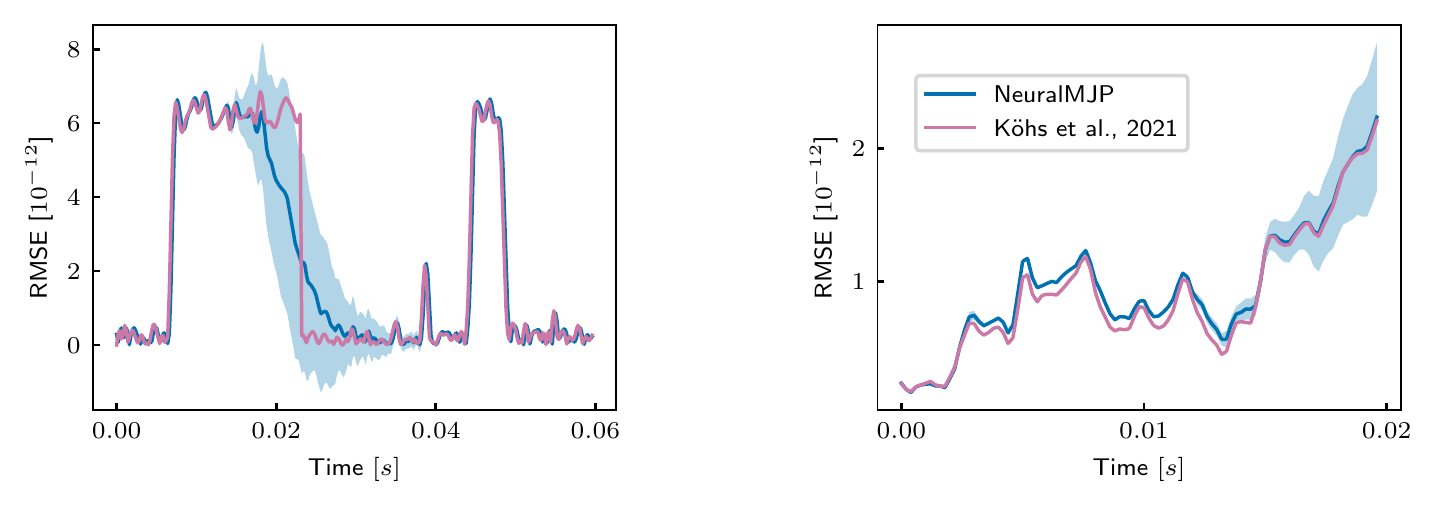}}{
    Comparison of prediction capabilities of two models trained on Switching Ion Channel data. Two subplots, containing RMSE for predictions on selected training data and on single test trajectory. The performance of both models is very similar.
    }
    \caption{Comparison of prediction capabilities of models trained on Switching Ion Channel data. Left: RMSE on observations after training set. Right: 25 time series extracted from training period. Posterior from \cite{koehs21} was extracted approximately from their Figure $4$.}
    \label{fig:ion_channel_prediction_comparison}
\end{figure}

\section{Switching Ion Channel Data}
\label{app:ion_channel}
\subsection{Data Description and Pre-Processing}
The ion permeability of the vial potassium channel $\text{Kcv}_{\tiny \text{MT325}}$ is experimentally  measured by applying a voltage drop across the cell membrane and recording the ion flow. 
The current has been measured with a frequency of $5$ kHz and a voltage of $140$ mV for a total of $34.323$ seconds or $171615$ observations. 

\citet{koehs21} consider the one-second time window from observation $9200$ to $14200$. 
We cut this time window into $50$ time series of length $100$, create $5$ batches of size $10$ and use all of them for training. 
Additional test and validation time series of length $100$ are taken from parts of the trajectory outside of this time window. 
 
The observation time for the first observation in each time series is set to $0$ and subsequent observation times are adjusted accordingly, retaining the measurement frequency from the original trajectory.

The time series for prediction includes observations from $14100$ to $14500$, where the first $100$ observations are used to infer the posterior distribution at observation $14200$. 

We also experiment with the full dataset. 
After removing a burn-in time of $5000$ observations from the start of the measurements, we split the trajectory into prediction, train, test and validation sets analogous to Section \ref{appendix:datadescription:adp}. 

We use $100$ time series of length $200$ for prediction and $10$ time series of length $100$ for test and validation each. 
This leaves $1626$ time series of length $100$ for training, from which we create $25$ batches of size $64$ and a additional batch with the remaining time series.

We normalize observation times and values to lie in $[0,1]$ when training on both datasets.

\begin{table}[b]
\caption{Mean first-passage times, expressed in seconds, of models trained on Switching Ion Channel one-second window data and the full dataset. Comparing available results from \cite{koehs21} and NeuralMJP. Entry $j$ in row $i$ is mean first-passage time of transition $i \rightarrow j$ of the corresponding model. }
\label{tab:ion_mfpt_appendix}
\vskip 0.15in
\begin{center}
\begin{small}
\begin{sc}
\begin{tabular}{r|ccc|ccc|ccc}
\toprule
            &  \multicolumn{6}{c|}{One-Second Window}  & \multicolumn{3}{c}{Full Dataset} \\
            & \multicolumn{3}{c|}{\citet{koehs21}} & \multicolumn{3}{c|}{NeuralMJP} &   \multicolumn{3}{c}{NeuralMJP}   \\
    $\tau_{ij} / s$ & Bottom & Middle & Top       & Bottom & Middle & Top         &   Bottom & Middle & Top   \\
    \midrule                                                                     
        Bottom & $0.   $ & $0.068$ & $0.054$   &   $0.   $ & $0.019$ & $0.031$    &  $0.   $ & $0.008$ & $0.014$   \\ 
        Middle & $0.133$ & $0.   $ & $0.033$   &   $0.083$ & $0.   $ & $0.014$    &  $0.031$ & $0.   $ & $0.006$   \\  
        Top    & $0.181$ & $0.092$ & $0.   $   &   $0.119$ & $0.038$ & $0.   $    &  $0.048$ & $0.018$ & $0.   $   \\    
\end{tabular}
\end{sc}
\end{small}
\end{center}
\vskip -0.1in
\label{tab:ion_channel_mfps_comparison}
\end{table}

\newpage
\subsection{Modelling and Results}
We train three-state NeuralMJPs with gaussian emission models on both both datasets.
They recover the three known configurations of the ion channel. 

We compare our results to the results of \citet{koehs21} on the one-second window dataset. 

The learned dynamics of both models, here represented by mean first-passage times in Table~\ref{tab:ion_channel_mfps_comparison}, differ by their resolution of transitions between \textsc{Top} and \textsc{Bottom}. 
NeuralMJP resolves them by short transitions to the \textsc{Middle} state, indicated by the short mean first-passage times to and from state \textsc{Middle}. 
These are much longer for \citet{koehs21}, indicating that transitions between \textsc{Top} and \textsc{Bottom} do not pass through state \textsc{Middle}.

This structural difference is also reflected in the significantly shorter learned time scales (see Table~\ref{tab:ion_channel_time_scales_comparison}). 

Figure~\ref{fig:ion_channel_prediction_comparison} shows that the prediction capabilities of both models are very similar, despite their structural differences in the learned dynamics. 

Moreover, Table~\ref{tab:ion_channel_time_scales_comparison} reveals that NeuralMJP learns different relaxation time scales for each dataset. 
It captures a data shift over the complete trajectory, where the \textsc{Middle} configuration is only visited for very short periods and less frequently.
%

The fact that the relaxation time scales are shorter in the \textsc{Full} dataset, as opposite to the \textsc{1 sec} dataset, reflect the fact that the time-averaged pdf describing the system of the \textsc{Full} dataset is closer to the stationary distribution, than the corresponding time-averaged pdf of the \textsc{1 sec} dataset.

In other words, this result simply reflects the ergodicity of the system.

\begin{table}[t]
\caption{Ordered relaxation time scales in seconds of models trained on Switching Ion Channel data.}
\label{tab:ion_timescales_appendix}
\vskip 0.15in
\begin{center}
\begin{small}
\begin{sc}
\begin{tabular}{rcc}
\toprule
                                & \multicolumn{2}{c}{Ordered time scales} \\
\midrule
         \citet{koehs21} &        $0.01228  $ &     $0.03396  $ \\
\midrule
                     NeuralMJP ($1$ sec) &    $0.00295$  & $0.02116  $ \\
                     NeuralMJP (full) &      $0.00099 $ &  $0.0097$ \\
\bottomrule
\end{tabular}
\end{sc}
\end{small}
\end{center}
\label{tab:ion_channel_time_scales_comparison}
\vskip -0.1in
\end{table}

\begin{figure*}[b!]
    \centering
    \pdftooltip{\includegraphics{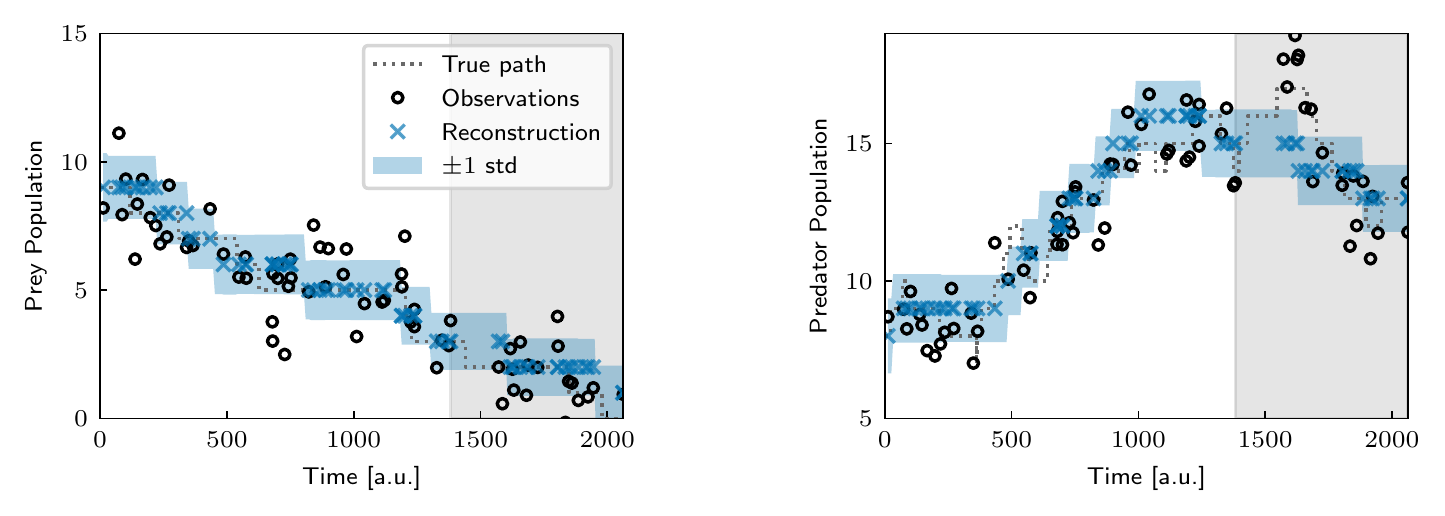}}{
    Lotka-Volterra trajectory with observations on a irregular grid, split into two subplots for predator and prey population. The prey population declines steadily from around $10$ to $0$. The predator population first increases from around $10$ to $15$ and then starts to decline again. Each subplot is split into a observation and prediction window. NeuralMJP reconstructions capture the overall trend of the trajectory closely in both windows.
    }
    \caption{Lotka-Volterra trajectory from the \textsc{Irregular Grid} dataset, with the true underlying path (black dots) and the noisy observations (black circles). The white background represents the test part of the trajectory, the grey shaded background on the right represents the prediction part.  Reconstruction of NeuralMJP (blue crosses) with learned standard deviation of the emission model (blue shaded) follows the true path closely on the test and prediction parts of the trajectory.}
    \label{fig:lv_prediction}
\end{figure*}

\newpage

\section{Lotka-Volterra}
\label{appendix:datadescription:lv}
\subsection{Data Description and Pre-Processing}
The Lotka-Volterra (LV) process is a MJP on $\mathbb{N}\times\mathbb{N}$ that models the coupled population levels of prey and predator species over time. 
As a pure birth-death process, only transitions to neighbouring states are possible. 
Moreover, it is assumed that population levels of both species can not change at the exact same time.

To describe the transition rates, let $x \in \mathbb{N}$ and $y \in \mathbb{N}$ denote the population levels of prey and predator respectively. 
The potentially non-zero transition rates of the Lotka-Volterra process are
\begin{align*}
    f(x+1, y \mid x, y) &= \alpha x         &f(x-1,y \mid x,y) &= \beta xy \\
    f(x, y+1 \mid x, y) &= \delta xy        &f(x,y-1 \mid x,y) &= \gamma y
\end{align*}
where $\alpha, \beta, \delta, \gamma \in \mathbb{R}^+$ are parameters of the model.

We sample trajectories of LV with parameters $\alpha = 0.0005$, $\beta = 0.0001$, $\delta=0.0001$ and $\gamma = 0.0005$ and a time window of $[0, 1500]$. 
Like \citet{opper07}, we add a small constant $10^{-6}$ to the transition rates $f(1,y \mid 0, y)$ and $f(x,1 \mid x, 0)$, which would be zero otherwise. 
Initial population levels are sampled uniformly random from $[5,20]\times[5,20]$.

As in Appendix \ref{app:dfr}, we sample $5000$ LV trajectories and observe them at three observation grids of length $50$. 
We use $70$ batches of size $64$ for training and leave the remaining time series for test and validation. 
For prediction, we generate $100$ additional time series of length $100$, just as in Appendix \ref{app:dfr}. 
The observed population levels are corrupted with samples from $\mathcal{N}(0,1)$. 
These observations are loosely bounded from above by $60$. 

\begin{table}[t]
\caption{LV parameters learned by NeuralMJP on multiple datasets. Additionally, KL divergence and reconstruction loss on the test set. Values are reported as mean and standard deviations of $5$ runs of NeuralMJP. Parameter values are reported in $10^{-4}$.}
\label{tab:app-lv-extra-params}
\vskip 0.1in
\begin{center}
\begin{small}
\begin{sc}
\begin{tabular}{rcccccc}
\toprule
                           &        KL & NLL &          $\alpha$ &                   $\beta$ &                  $\delta$ &                  $\gamma$ \\
\midrule
                        Ground Truth &      -  &- &       $5$ &               $1$ &               $1$ &               $5$ \\
                        \citet{opper07} &   -   & - &       $13.5$ &               $2.32$ &               $1.78$ &               $15.7$ \\
\midrule
                 Irregular Grid & $31 \pm 2$     & $166 \pm 2$ & $4.3 \pm 0.6$ & $0.94 \pm 0.07$ & $ 0.83 \pm 0.09$ & $ 4.1 \pm 0.4$ \\
                    Shared Grid & $36 \pm 4$     & $155 \pm 2$ & $5.4 \pm 0.6$ & $1.10 \pm 0.07$ & $ 1.00 \pm 0.06$ & $ 4.9 \pm 0.4$ \\
                   Regular Grid & $28 \pm 5$     & $163 \pm 4$ & $4.5 \pm 0.6$ & $1.00 \pm 0.03$ & $ 0.90 \pm 0.08$ & $ 4.3 \pm 0.5$ \\
             Single Trajectory  &            $192 \pm 114$  & $13  \pm 24$ &  $2.1 \pm 0.6$ & $0.67 \pm 0.06$ & $ 1.05 \pm 0.32$ & $ 9.3 \pm 8.1$ \\
\bottomrule
\end{tabular}
\end{sc}
\end{small}
\end{center}
\vskip -0.1in
\end{table}

\subsection{Modelling and Results}
We follow \citet{opper07} and use a mean-field approach by training NeuralMJP with two posterior MJPs with $60$ states each. 
Each MJP is a birth-death process and transitions out of lower and upper bounds are impossible. 

The prior is constrained to a LV process and parametrized as above, but transition rates for transitions out of the boundary are set to $0$. 
We use a gaussian emission model, where the sampled latent state are the gaussian mean and only the diagonal covariance matrix is learned. 
Thus, each species is linked to one posterior process, which is in accordance with \cite{opper07}.

We summarize the losses of our models on test sets and the inferred LV parameters in Table \ref{tab:app-lv-extra-params} and the RMSE on a prediction set in Table \ref{tab:app-lv-prediction}. 
The inferred parameters are close to the ground-truth values and consistent across multiple runs of NeuralMJP with different initializations. 

In Figure \ref{fig:lv_prediction} we see that the posterior MJP follows the underlying dynamics closely and learns a reasonable standard deviation.  
Moreover, prediction with the prior process captures the trend of the continued sample path.

\begin{table}[t!]
\caption{Reconstruction RMSE in test and prediction sets at a single observation for NeuralMJPs trained on LV datasets. For prediction, we consider the RMSE at the $1$st, $5$th and $10$th observation in the prediction set. Values are reported as mean and standard deviation of $5$ runs of NeuralMJP. Values for \citet{opper07} are calculated based on their Figure 1.}
\label{tab:app-lv-prediction}
\vskip 0.1in
\begin{center}
\begin{small}
\begin{sc}
\begin{tabular}{rcccc}
\toprule
                           &         RMSE Test & \multicolumn{3}{c}{RMSE Prediction at Observation} \\
                                     &           &                 1 &                 5 &                 10 \\
\midrule
                     \citet{opper07} &   $2.672  $ &   $3.000  $ &   $9.487  $ &   $11.402  $ \\
\midrule
                 Irregular Grid & $1.87 \pm 0.04$ & $2.67 \pm 0.17$ & $3.1 \pm 0.1$ &  $3.3 \pm 0.1$ \\
                    Shared Grid & $1.67 \pm 0.04$ & $2.33 \pm 0.08$ & $2.8 \pm 0.2$ &  $3.1 \pm 0.2$ \\
                   Regular Grid & $1.80 \pm 0.08$ & $1.75 \pm 0.14$ & $2.4 \pm 0.2$ &  $2.7 \pm 0.2$ \\
              Single Trajectory & $2.02 \pm 0.04$ & $5.69 \pm 2.66$ & $9.2 \pm 3.6$ & $10.6 \pm 4.3$ \\
\bottomrule
\end{tabular}
\end{sc}
\end{small}
\end{center}
\vskip -0.1in
\end{table}

\newpage

\subsection{Single Lotka-Volterra Time Series}
We also experiment with the single, short Lotka-Volterra time series from Figure $1$ of \citet{opper07}. 

They simulated the trajectory with parameters $\alpha, \beta, \delta, \gamma$ from above and chose initial states $x=19$ and $y=7$. 
Observations at $16$ equidistant times in $[0, 1500]$ were then corrupted by a two-sided, discrete exponential noise.

They also provide the continuation of the trajectory up to time $3000$. 
We use this continuation to create a prediction set by observing it at $15$ equidistant times in $[1600, 3000]$.
Note that we do not add any noise these additional observations. 

\begin{figure*}[b!]
    \centering
    \pdftooltip{\includegraphics{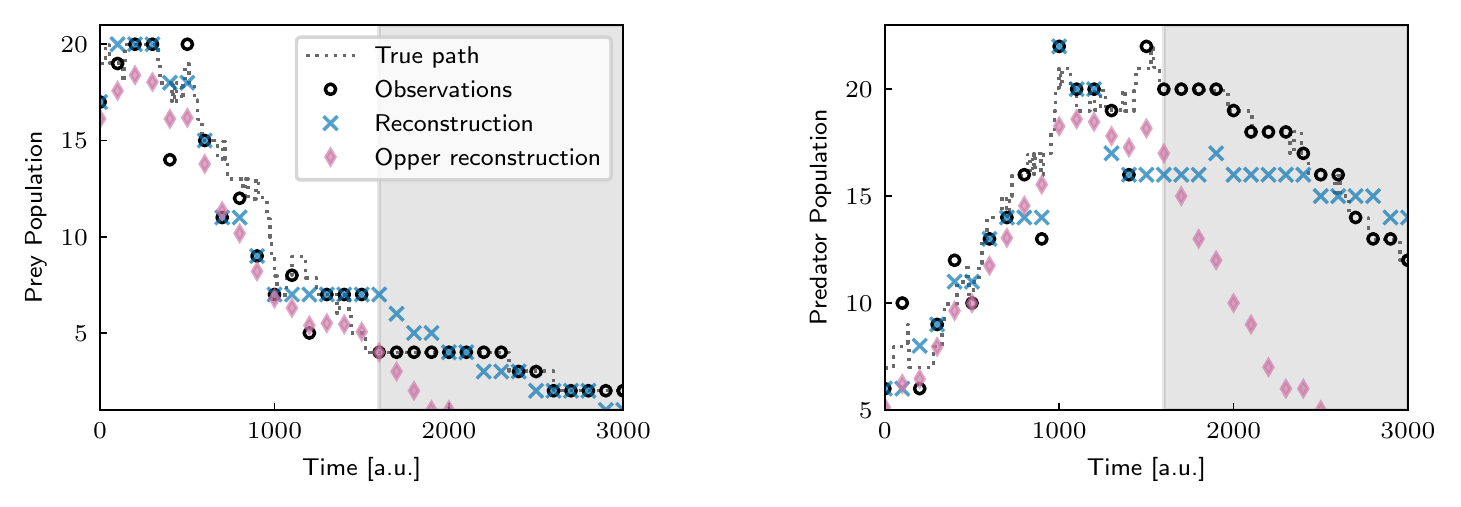}}{Lotka-Volterra trajectory from \cite{opper07}, split into two subplots for predator and prey population. The prey population declines steadily from around $20$ to $0$. The predator population first increases from around $5$ to $20$ and starts to decline afterwards. Each subplot is split into a observation and prediction window. Reconstructions from NeuralMJP and \cite{opper07} follow the trajectory closely in the observation window. In the prediction window, NeuralMJP reconstructs the observations more accurately.}
    \caption{Lotka-Volterra single time series from \cite{opper07}. The training part of the trajectory is on the white background to the left, while the prediction part of the trajectory is on the grey background to the right. Reconstruction and prediction with NeuralMJP trained on this single trajectory (blue crosses). Comparison to the results from \cite{opper07} (red diamonds), where their learned master equation was solved for prediction, with initial value approximated from Figure 1 \cite{opper07}.}
    \label{fig:lv_prediction-2}
\end{figure*}

We train NeuralMJP with the setup from above on this single trajectory. 
The results in Table \ref{tab:app-lv-extra-params} show that our model predicts parameters closer to the ground truth, compared to \citet{opper07}, although there is a lot of variance in $\gamma$.  
This translates to more accurate predictions, as depicted in Figure \ref{fig:lv_prediction-2}.

\newpage

\begin{figure}[h!]
    \centering
    \pdftooltip{\includegraphics[scale=0.2]{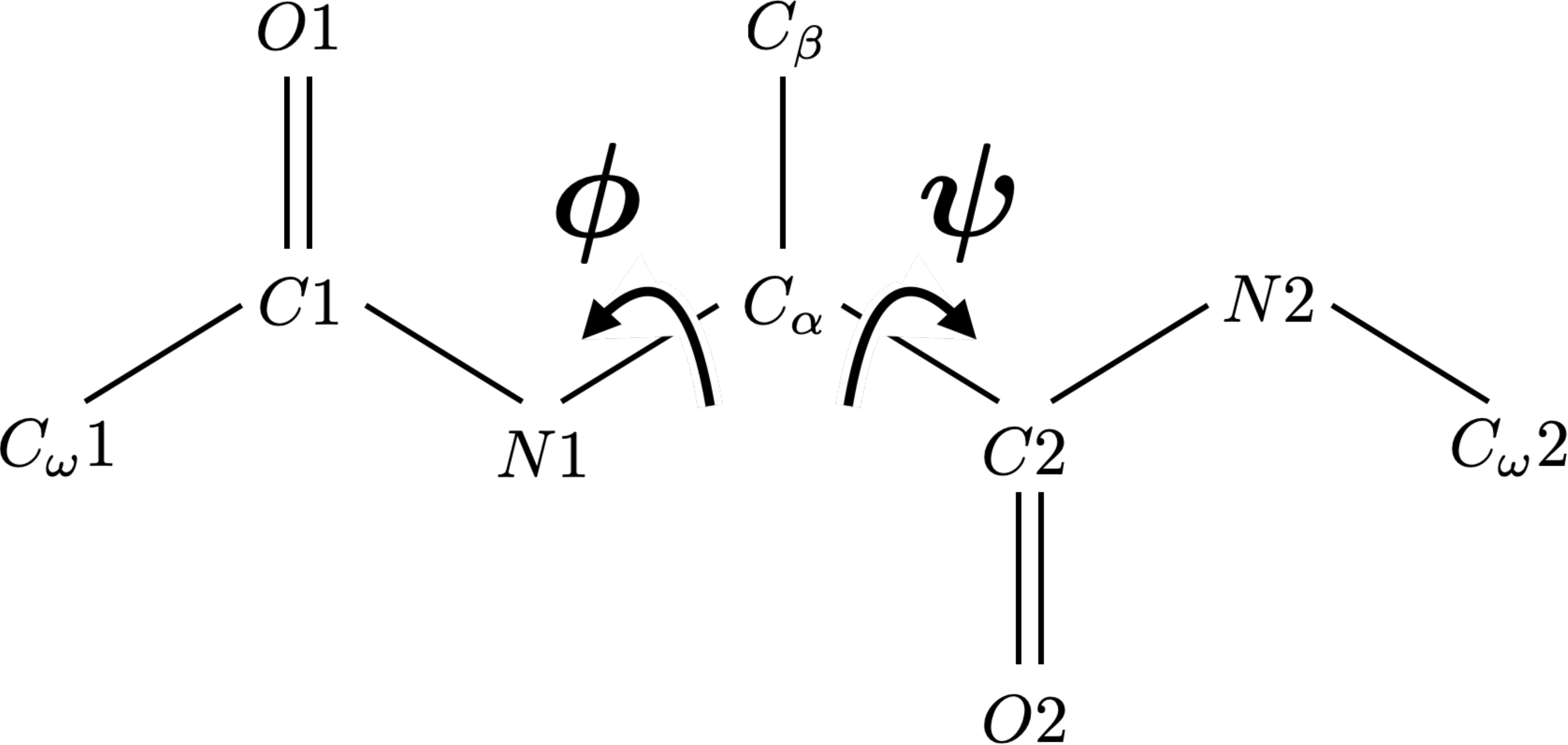}}{
    Molecular model of the heavy atoms of the Alanine Dipeptide Protein. 
    }
    \vspace{-0.1cm}
    \caption{Heavy atoms of the Alanine Dipeptide Protein with locations of Ramachandran angles $\psi$ and $\phi$.}
    \label{fig:adp_heavy_atoms}
\end{figure}

\section{Alanine Dipeptide Protein}
\label{appendix:datadescription:adp}
\subsection{Data Description and Pre-Processing}

The Alanine Dipeptide Protein (ADP) data used in Section \ref{sec:adp} is a $10^6$ step all atoms molecular dynamics simulation with stepsize $1$ picosecond (ps). 
The system is in equilibrium and the data has been aligned to the first frame, removing global translations and rotations of the protein. 

We first transform the observations into Ramachandran angles, which are dihedral angles of two backbone atom sequences. 
The dihedral angles for atom sequences $N1-C_\alpha-C2-N2$ and $C1-N1-C_\alpha-C2$ are denoted by $\psi$ and $\phi$ respectively. 
Their locations in the atom are depicted in Figure \ref{fig:adp_heavy_atoms}. 
Partially following \cite{varolguenes19}, we use $\sin \psi$, $\cos \psi$, $\sin \phi$ and $\cos \phi$ as features, which take the periodicity of the angles into consideration. 

For prediction, we first extract $100$ non-overlapping time series of length $200$ from this trajectory. 
Afterwards, the remaining trajectory is cut up into time series of length $100$. 
We create $140$ batches of size $64$ for training and leave the other time series for test and validation. 

The observation time for the first observation in each time series is set to $0$ and subsequent observation times are adjusted accordingly, retaining the stepsize from the original trajectory. 
We normalize observation times to lie in $[0,1]$.

\begin{table}[b]
    \caption{Mean first-passage times in nanoseconds of the six-state NeuralMJP trained on APD data with standard deviation from sampling the generative prior. These results are depicted in Table \ref{tab:adp_mfps} without standard deviation. Entry $j$ in row $i$ is the mean first-passage time of transitions $i \rightarrow j$.}
    \vskip 0.15in
\begin{center}
\begin{small}
\begin{sc}
    \begin{tabular}{crrrrrr}
    \toprule
        $\tau_{ij} / ns$ & $\rom{1}$ & $\rom{2}$ & $\rom{3}$ & $\rom{4}$ & $\rom{5}$ & $\rom{6}$ \\
    \midrule
        $\rom{1}$ & $0.000 \pm 0.000$   & $0.028 \pm 0.002$     & $0.290 \pm 0.030$     & $0.107 \pm 0.006$     & $14.000 \pm 4.000$    & $16.000 \pm 4.000$\\
        $\rom{2}$ & $0.032 \pm 0.003$   & $0.000 \pm 0.000$     & $0.290 \pm 0.030$     & $0.103 \pm 0.006$     & $14.000 \pm 4.000$    & $16.000 \pm 4.000$\\
        $\rom{3}$ & $0.130 \pm 0.010$   & $0.130 \pm 0.010$     & $0.000 \pm 0.000$     & $0.063 \pm 0.005$     & $14.000 \pm 4.000$    & $16.000 \pm 4.000$\\
        $\rom{4}$ & $0.073 \pm 0.008$   & $0.070 \pm 0.010$     & $0.190 \pm 0.020$     & $0.000 \pm 0.000$     & $14.000 \pm 4.000$    & $16.000 \pm 4.000$\\
        $\rom{5}$ & $0.920 \pm 0.220$   & $0.910 \pm 0.220$     & $1.100 \pm 0.240$     & $0.900 \pm 0.200$     & $0.000  \pm 0.000$    & $4.000  \pm 1.000$\\
        $\rom{6}$ & $0.800 \pm 0.210$   & $0.800 \pm 0.210$     & $0.950 \pm 0.210$     & $0.800 \pm 0.200$     & $3.000  \pm 1.000$    & $0.000  \pm 0.000$\\
    \bottomrule
    \end{tabular}
    \end{sc}
\end{small}
\end{center}
\label{tab:adp_first_passage_appendix}
\vskip -0.1in
\end{table}

\newpage

\begin{table}[t]
    \caption{Transition rates from the six-state NeuralMJP trained on ADP data. They are expressed in nanoseconds with standard deviation from sampling the generative prior. Entry $j$ in row $i$ is the rate of transition $i \rightarrow j$.}
        \vskip 0.15in
\begin{center}
\begin{small}
\begin{sc}
    \begin{tabular}{ccccccc}
    \toprule
    $i \rightarrow j$ & $\rom{1}$ & $\rom{2}$ & $\rom{3}$ & $\rom{4}$ & $\rom{5}$ & $\rom{6}$ \\
    \midrule
   $\rom{1}$ & $0.00  \pm 0.00 $   & $53.00 \pm 4.00$   & $0.19  \pm 0.06 $ & $7.90  \pm 0.90 $    & $0.06  \pm 0.03$ & $0.02 \pm 0.01 $\\
   $\rom{2}$ & $47.00 \pm 3.00 $   & $0.00  \pm 0.00$   & $0.05  \pm 0.03 $ & $12.00 \pm 0.90 $    & $0.04  \pm 0.02$ & $0.01 \pm 0.01 $\\
   $\rom{3}$ & $0.28  \pm 0.09 $   & $0.13  \pm 0.05$   & $0.00  \pm 0.00 $ & $17.00 \pm 1.40 $    & $0.02  \pm 0.01$ & $0.04 \pm 0.02 $\\
   $\rom{4}$ & $36.00 \pm 3.00 $   & $26.90 \pm 2.70$   & $41.00 \pm 4.00 $ & $0.00  \pm 0.00 $    & $0.30  \pm 0.10$ & $0.01 \pm 0.01 $\\
   $\rom{5}$ & $0.17  \pm 0.06 $   & $0.20  \pm 0.10$   & $0.40  \pm 0.20 $ & $0.20  \pm 0.20 $    & $0.00  \pm 0.00$ & $3.00 \pm 1.00 $\\
   $\rom{6}$ & $1.20  \pm 0.60 $   & $1.80  \pm 0.80$   & $0.50  \pm 0.20 $ & $0.70  \pm 0.40 $    & $19.00  \pm 7.50$& $0.00 \pm 0.00   $ \\
    \bottomrule
    \end{tabular}
    \end{sc}
\end{small}
\end{center}
\label{tab:adp_trans_rates_appendix}
\vskip -0.1in
\end{table}

\subsection{Modelling and Results}
We train NeuralMJPs with six, three and two states and unconstrained prior and posterior MJPs. 
For this dataset, we chose a gaussian emission model and learn the gaussian mean and the complete covariance matrix. 

\begin{figure}[b!]
\footnotesize
\begin{floatrow}
    \ffigbox[0.48\textwidth]{
       \pdftooltip{\includegraphics[width=0.49\textwidth]{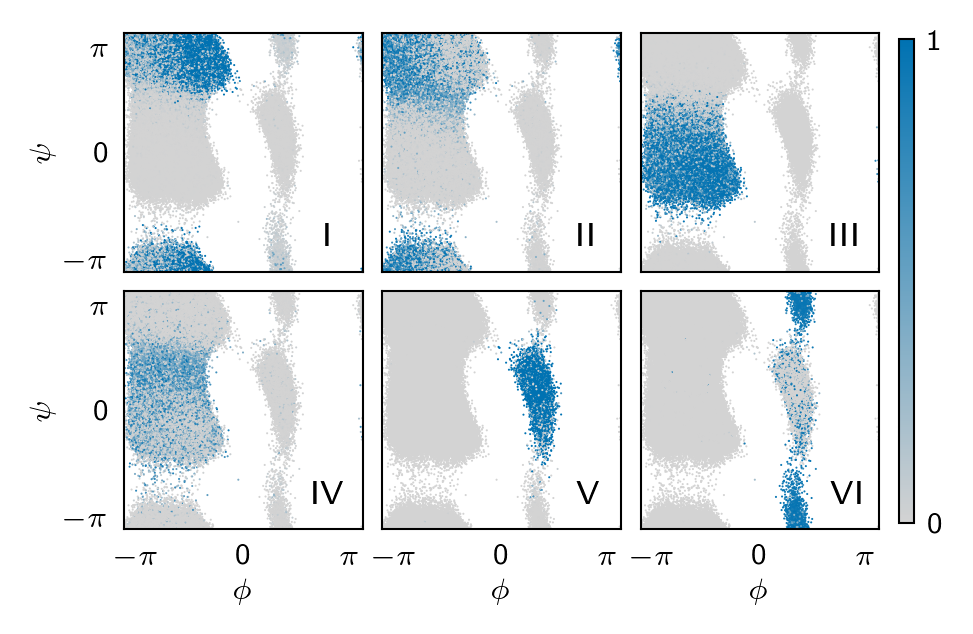}}{State separation of another run of a six-state NeuralMJP on Alanine Dipeptide Protein data. Similar to Figure 4, but clusters are less well defined and have a slightly different shape.}
    }{\vspace{-0.3cm}
        \caption{ADP state separation from another six-state NeuralMJP run that differs from t he one presented in Figure \ref{fig:adp_6_states}. For each observation, the intensity of blue in subplot $i= \rom{1}, \dots, \rom{6}$ indicates the posterior probability mass for state $i$ at that observation. }
        \label{fig:alt_6_state_separation}
    }
    
    \ffigbox[0.48\textwidth]{
        \pdftooltip{\includegraphics[width=0.49\textwidth]{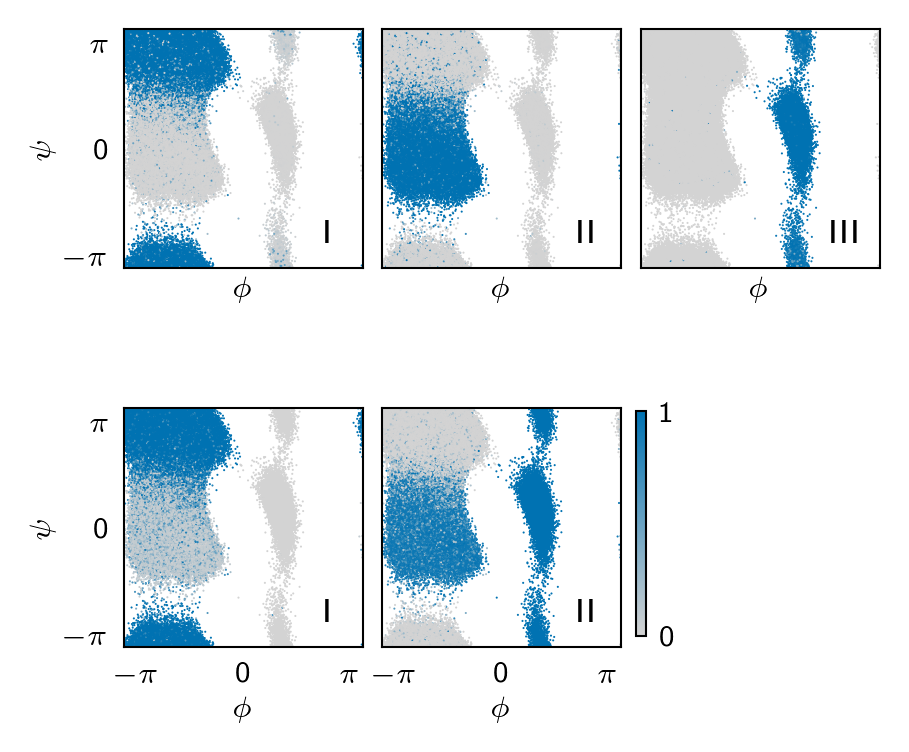}}{
    State separation of a three-state and a two-state NeuralMJP model trained on Alanine Dipeptide Protein data. There are either two or three subplots, each containing all observations of the dataset represented in Ramachandran angles. The coloring of a datapoint in a subplot indicates the posterior probability mass that NeuralMJP puts on the datapoint for each state. There are two or three distinct clusters, depending on the model. At their edges, the cluster blend together, indicating a fuzzy separation of states learned by NeuralMJP.
    }
    }{\vspace{-0.3cm}
        \caption{State separation of three and two-state NeuralMJPs trained on ADP data. For each observation, the intensity of blue in subplot $i= \rom{1}, \rom{2}, \rom{3}$ indicates the posterior probability mass for state $i$ at that observation. }
        \label{fig:adp_2_3_states}
    }
\end{floatrow}
\end{figure}

The separation of states in models with two and three states is depicted in Figure~\ref{fig:adp_2_3_states}. 
They resemble two slower relaxation time scales of the protein. 

Table~\ref{tab:adp_first_passage_appendix} contains the mean first-passage times of the six-state model presented in Section~\ref{sec:adp}. 
Note that it contains the values from Table \ref{tab:adp_mfps}, with additional standard deviation from sampling the prior rates. 
The learned transition rates of that model are in Table \ref{tab:adp_trans_rates_appendix}. 
We remark that the results for that model were discussed thoroughly in Section~\ref{sec:adp}.

\newpage

\subsection{Model Instabilities}
\label{appendix:adp_instabilities}
Results across multiple runs of six-state NeuralMJPs on the ADP data are not as stable as the results of multiple runs on other experiments. 

One source of instability is the failure of separating state $\rom{6}$ from state $\rom{5}$, which can be attributed to its rare occurrence in the data. 
The same failure has been reported by \citet{mardt17} and \citet{varolguenes19}, who train end-to-end, unsupervised Markov models on similar ADP trajectories. 

Note that this failure cannot be easily detected based on training dynamics or metrics. 
In Table~\ref{app:time scales_15_adp}, we display the results of $15$ runs of NeuralMJP with six states on the ADP data. 
Only a few of them resolve state $\rom{6}$ accurately. 
However, they are not immediately discernible from each other based on their reconstruction performance, here represented by RMSE, or learned dynamics, here represented by the learned time scales. 

Another source of instability is the size, shape and position of clusters learned by our model, in particular in the $\phi < 0$ region of the Ramachandran plane. 
By \citet{mironov19}, this region contains several low energy minima of the protein, which can be grouped by their proximity in the Ramachandran plane and other configuration properties.

However, unsupervised clustering of that region by NeuralMJP does not produce such consistent clusters. 
Compare Figures \ref{fig:adp_6_states} and \ref{fig:alt_6_state_separation} for two runs of six-state NeuralMJPs. 
Although both runs produce comparable results on the test set, there is a noticeable difference in their clusters. 
In particular, the edges of clusters in Figure~\ref{fig:alt_6_state_separation} are not as well defined as in Figure~\ref{fig:adp_6_states}. 

The posterior distribution depends significantly on the particular clustering, so these inconsistencies extend to differences in the learned dynamics. 
Therefore, comparing models by their transition rates or stationary distributions is only possible, if their clusters are comparable.

Finally, some prior rates exhibit a large sampling standard deviation, which can lead to considerable uncertainty of eigenvalues, relaxation times and mean first-passage times, even for a single run. 
See Table~\ref{tab:adp_trans_rates_appendix}, where some of the smaller rates display considerable standard deviation. 

Considering the above, we selected the run presented in Section \ref{sec:adp} based on its separation of states in the $\phi > 0$ region of the Ramachandran plane and its distinct clusters that we can identify with clusters found, for example, by \citet{mironov19} and \citet{varolguenes19}.

\begin{table}[b!]
\caption{RMSE and relaxation time scales of $15$ NeuralMJP with six states on ADP data. Standard deviation in time scales is due to sampling the generative prior. Standard deviation in RMSE due to the $100$ time series in test set. First $5$ experiments were handpicked by their separation of states $\rom{5}$ and $\rom{6}$. Out of the other $10$ non-handpicked experiments, only experiment $7$ separated states $\rom{5}$ and $\rom{6}.$ VAMPNets results are taken from \citet{mardt17}, GMVAE from \citet{varolguenes19}.}
    \vskip 0.15in
\begin{center}
\begin{small}
\begin{sc}
\begin{tabular}{lcccccc}
\toprule
 Experiment      & RMSE         &   \multicolumn{5}{c}{Ordered Time Scales in $ns$} \\
\midrule
 VAMPnets      &    -            & $0.008$ & $0.009$ & $0.055$ & $0.065$ & $1.920$ \\
 GMVAE &    -            & $0.003$ & $0.003$ & $0.033$ & $0.065$ & $1.430$ \\
\midrule

 1     & $0.6 \pm 0.1$   & $0.008\pm0.000$ & $0.011\pm0.001$ & $0.027\pm0.005$ & $0.084\pm0.005$ & $0.800\pm0.200$ \\
 2     & $0.7 \pm 0.2$   & $0.021\pm0.001$ & $0.048\pm0.004$ & $0.055\pm0.003$ & $0.500\pm0.100$ & $1.000\pm0.200$ \\
 3     & $0.7 \pm 0.1$   & $0.012\pm0.000$ & $0.014\pm0.001$ & $0.040\pm0.010$ & $0.068\pm0.004$ & $1.500\pm0.300$ \\
 4     & $0.6 \pm 0.1$   & $0.009\pm0.000$ & $0.013\pm0.001$ & $0.071\pm0.005$ & $0.250\pm0.030$ & $0.410\pm0.070$ \\
 5     & $0.6 \pm 0.1$   & $0.009\pm0.000$ & $0.009\pm0.000$ & $0.040\pm0.010$ & $0.069\pm0.004$ & $0.800\pm0.200$ \\
\midrule               
 6     & $0.6 \pm 0.1$   & $0.006\pm0.000$ & $0.008\pm0.000$ & $0.010\pm0.000$ & $0.059\pm0.001$ & $0.700\pm0.100$ \\
 7     & $0.7 \pm 0.1$   & $0.012\pm0.000$ & $0.012\pm0.000$ & $0.060\pm0.001$ & $0.080\pm0.004$ & $1.300\pm0.200$ \\
 8     & $0.6 \pm 0.1$   & $0.008\pm0.000$ & $0.010\pm0.000$ & $0.013\pm0.000$ & $0.073\pm0.001$ & $0.800\pm0.100$ \\
 9     & $0.7 \pm 0.1$   & $0.010\pm0.000$ & $0.015\pm0.000$ & $0.017\pm0.000$ & $0.064\pm0.001$ & $0.670\pm0.080$ \\
 10    & $0.6 \pm 0.1$   & $0.005\pm0.000$ & $0.009\pm0.000$ & $0.010\pm0.000$ & $0.077\pm0.001$ & $0.500\pm0.050$ \\
 11    & $0.7 \pm 0.2$   & $0.010\pm0.000$ & $0.011\pm0.000$ & $0.018\pm0.000$ & $0.053\pm0.001$ & $0.800\pm0.100$ \\
 12    & $0.7 \pm 0.2$   & $0.007\pm0.000$ & $0.009\pm0.000$ & $0.011\pm0.000$ & $0.063\pm0.001$ & $0.800\pm0.100$ \\
 13    & $0.6 \pm 0.1$   & $0.007\pm0.000$ & $0.008\pm0.000$ & $0.013\pm0.000$ & $0.060\pm0.000$ & $0.700\pm0.100$ \\
 14    & $0.6 \pm 0.1$   & $0.007\pm0.000$ & $0.008\pm0.000$ & $0.011\pm0.000$ & $0.063\pm0.001$ & $0.610\pm0.060$ \\
 15    & $0.6 \pm 0.1$   & $0.005\pm0.000$ & $0.006\pm0.000$ & $0.008\pm0.000$ & $0.070\pm0.000$ & $0.650\pm0.060$ \\
\bottomrule
\end{tabular}
\end{sc}
\end{small}
\end{center}
\label{app:time scales_15_adp}
\vskip -0.1in
\end{table}

\newpage
\section{A Simple Protein Folding Model}
\label{appendix:datadescription_bd}
\subsection{Data Description and Pre-Processing}
\citet{mardt17} simulate a $10^5$ time step trajectory of the $5$-dimensional Brownian dynamics
\begin{equation*}
    dx(t) = - \nabla U(x(t)) + \sqrt{2} dW(t) \quad ,
\end{equation*}
where $U(x)$ only depends on the norm $r(x) = \lvert x \rvert$ of $x$: 
\begin{equation*}
    U(x) = 
    \begin{cases}
        -2.5[r(x) - 3]^2    &\text{, if } r(x) < 3 \\
        0.5[r(x)-3]^3 - [r(x)-3]^2  &\text{, if } r(x) \geq 3
    \end{cases}
\end{equation*}

These dynamics are bistable along the norm $r(x)$ and serve them as a toy protein folding model, representing a folded and unfolded state.

We follow \citet{mardt17} and use the Euler method with $\Delta t=0.1$ to simulate trajectories of these dynamics. 
Initial states are sampled from $\mathcal{N}(-\mathbf{1}, 4 I)$, where $\mathbf{1} = [1,1,1,1,1] \in \mathbb{R}^5$ and $I \in \mathbb{R}^{5\times 5}$ is the identity matrix. 
The initial distribution was extracted from the code that accompanies their paper. 
Its samples are likely to be close to one of the bistable regions of the dynamics.

We simulate $1000$ trajectories with $100$ steps after a burn-in period of $1000$ steps, for a total of $10^5$ observations. 
During training, we normalize observation times to lie in $[0,1]$.
For training, we use $28$ batches of size $32$ and leave the remaining time series for test and validation.

We simulate $100$ additional time series for prediction, each with $200$ time steps after a burn-in period of $1000$ steps. 

\subsection{Modelling and Results}
We trained a NeuralMJP with two-state, unconstrained prior and posterior MJPs to capture the two stable regions along $r(x)$. 
For this dataset, we chose a gaussian emission model and learn the gaussian mean and diagonal covariance matrix. 

Indeed, NeuralMJP discerns the two stable regions. 
It learns a mean reconstruction close to $\*0$ for both states, but two different covariance matrices, one with smaller diagonal entries, representing the folded state, the other with larger diagonal entries, representing the unfolded state. 

The long-term dynamics, represented by the stationary distribution, of our continuous time model agrees with the discrete time model of \citet{mardt17}, see Table \ref{tab:app-tpf-stat-dist}. 
We provide the learned transition rates of our model in Table \ref{tab:app-tpf-trans-rates}.

\begin{figure}[t!]
\footnotesize
\newfloatcommand{capbtabbox}{table}[][]
\begin{floatrow}
    \capbtabbox{
        \begin{center}
        \begin{small}
        \begin{sc}
        \begin{tabular}{rcc}
        \toprule
             &            Low STD &       High STD \\
        \midrule
          \citet{mardt17} &                $0.73$ &            $0.27$ \\
        \midrule
              NeuralMJP &     $0.74 \pm 0.02$ & $0.26 \pm 0.02$ \\
        \bottomrule
        \end{tabular}
        \end{sc}
        \end{small}
        \end{center}
    }{\vspace{-0.3cm}
        \caption{Stationary distributions of models trained on the  dataset. Values of NeuralMJP are mean and standard deviation over $5$ runs. }
        \label{tab:app-tpf-stat-dist}
    }
    \capbtabbox{
        \begin{center}
        \begin{small}
        \begin{sc}
            \begin{tabular}{rc}
            \toprule
                        &  NeuralMJP \\
            \midrule
               Low STD $\rightarrow$ High STD &  $ 0.028 \pm 0.007 $          \\
               High STD $\rightarrow$ Low STD &    $0.085 \pm 0.011$          \\
            \bottomrule
            \end{tabular}
        \end{sc}
        \end{small}
        \end{center}
    }{\vspace{-0.3cm}
        \caption{Transition rates learned by NeuralMJP on the  dataset. Values are mean and standard deviation over $5$ runs.}
        \label{tab:app-tpf-trans-rates}
    }
\end{floatrow}
\end{figure}

\begin{table}[b!]
    \vspace{-0.2cm}\caption{Two-Mode Hybrid System transition rates including standard deviation over $5$ runs. Note that one run of NeuralMJP suffered from index collapse (see Appendix \ref{app:index_collapse}). Learned transition rates from the remaining runs are close to the ground truth. }
            \vskip 0.1in
\begin{center}
\begin{small}
\begin{sc}
    \begin{tabular}{ccc}
    \toprule
        & Bottom $\rightarrow$ Top &  Top $\rightarrow$ Bottom \\
    \midrule
    Ground Truth & $0.2$ & $0.2$ \\
    \citet{koehs21} & $0.64$  & $0.63$ \\
    \midrule
    NeuralMJP, 5 runs & $0.16 \pm 0.02$ & $0.90 \pm 0.90$ \\
    NeuralMJP, 4 runs & $0.19 \pm 0.02$ & $0.36 \pm 0.06 $ \\
    \bottomrule
\end{tabular}
\end{sc}
\end{small}
\end{center}
\vskip -0.1in
\label{tab:hybrid_switching_dynamics}
\end{table}

\newpage

\section{A Toy Two-Mode Switching System}
\label{appendix:datadescription:hybrid}
\subsection{Data Description and Pre-Processing}
\citet{koehs21} generate a time series of length $67$ from a trajectory of a switching stochastic differential equation with two modes. 
We generate this time series with the code provided by the authors. 
Here, we only provide a rough outline of the generation process for convenience and refer to \cite{koehs21} for a detailed description.

They consider a two-state, time-homogeneous MJP with transition rates
\begin{equation*}
    \begin{pmatrix}
    0    & 0.2 \\
    0.2  & 0
    \end{pmatrix} 
\end{equation*}

as the underlying switching process of their dynamics. 
A trajectory $z$ of this process defines a switching stochastic differential equation 
\begin{equation*}
    dy(t) = \alpha_{z(t)} (\beta_{z(t)} - y(t)) + 0.5 dW(t)
\end{equation*}
with $\alpha_1 = \alpha_2 = 1.5$, $\beta_1 = -1$ and $\beta_2 = 1$. 
This is understood as a system that switches between two Ornstein-Uhlenbeck processes with different drift terms, based on the state of the underlying switching process $z$.

\citet{koehs21} simulate a single trajectory of these dynamics, observe it at observation times sampled from a Poisson point process with intensity $\frac{1}{\lambda} = 0.35$ and corrupt the observations with samples from $\mathcal{N}(0, 0.1)$. 

During training, we normalize observation times and values to lie in $[0,1]$.

\subsection{Modelling and Results}
We train a two-state NeuralMJP with a gaussian emission model and learn the gaussian mean and standard deviation. 

\begin{figure*}[h]
    \centering
    \pdftooltip{\includegraphics{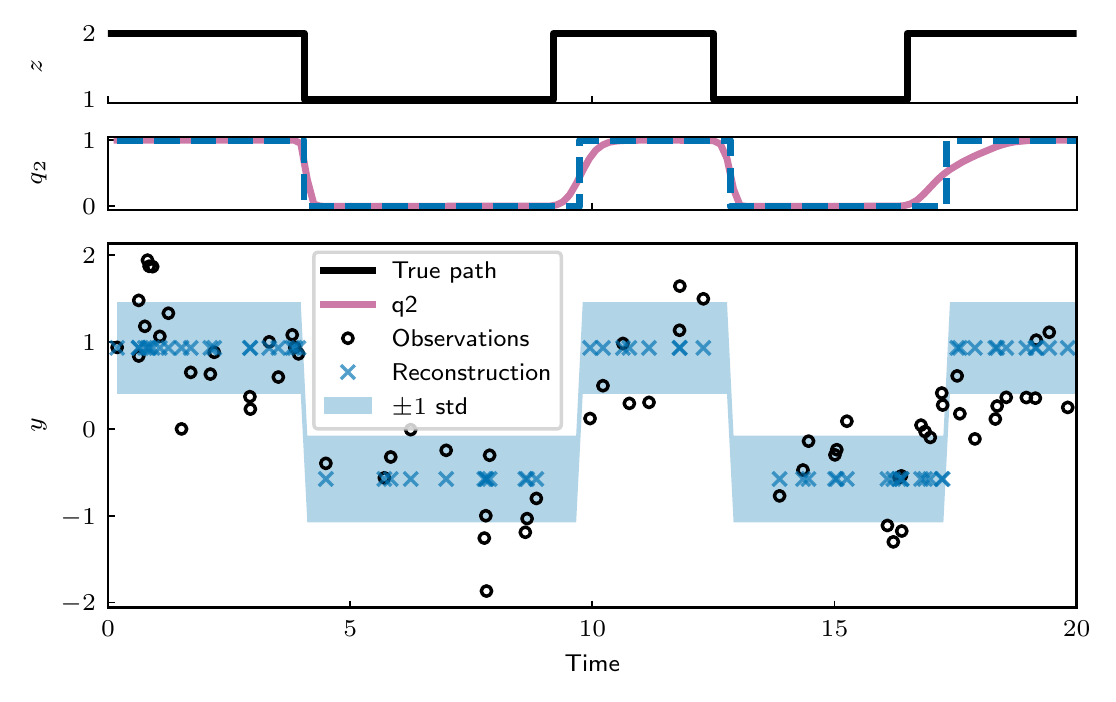}}{Synthetic Two-Mode Hybrid System trajectory, displayed in three subplots. Top subplot contains true underlying switching between two modes. Middle subplot contains reconstructed switching behaviour from NeuralMJP. Bottom subplot contains observations from the full system and reconstructions from NeuralMJP. NeuralMJP approximates the true switching dynamics closely and reconstructs the two modes of the full system well. }
    \caption{Results on Two-Mode Hybrid System. Bottom: Observations of the Hybrid System and reconstruction by NeuralMJP. Middle: Posterior probability of latent state corresponding to upper Ornstein-Uhlenbeck process (red solid line) and greedy prediction of that state (blue dashed line). Top: True underlying switching dynamics.}
    \label{fig:hybrid_switching_reconst}
\end{figure*}

NeuralMJP reconstructs observations stemming from both processes closely, see Figure \ref{fig:hybrid_switching_reconst}. 
Moreover, the posterior process approximates the true switching dynamics of the observed time series almost perfectly. 
Learned transition rates are closer to the ground-truth switching process than the model of \citet{koehs21} (see Table \ref{tab:hybrid_switching_dynamics}).

\end{document}